\documentclass[times,twocolumn,final]{elsarticle}
\usepackage{cnf}
\usepackage{framed,multirow}

\usepackage{amssymb}
\usepackage{latexsym}
\usepackage{placeins}
\usepackage{url}
\usepackage{xcolor}
\definecolor{newcolor}{rgb}{.8,.349,.1}
\usepackage{booktabs}
\usepackage{hyperref}
\usepackage{amsmath}
\usepackage{overpic}
\usepackage[switch,pagewise]{lineno} %Required by command \linenumbers below

\journal{Combustion and Flame}

\begin{document}

\verso{Zolfaghari et al.}

\begin{frontmatter}

\title{Hierarchical Multi-Fidelity Learning for Predicting Three-Dimensional Flame Wrinkling and Turbulent Burning Velocity}

\author[1]{Saghar {Zolfaghari}}

\author[2]{Yu {Xie}}

\author[2]{Junfeng {Yang}}

\author[1]{Safa {Jamali}\corref{cor1}}
\cortext[cor1]{Corresponding author: Department of Mechanical and Industrial Engineering, Northeastern University, Boston, MA 02115, USA.}
\emailauthor{s.jamali@northeastern.edu}{Safa Jamali}

\address[1]{Department of Mechanical and Industrial Engineering, Northeastern University, Boston, MA 02115, USA}

\address[2]{School of Mechanical Engineering, University of Leeds, Leeds, UK}

\begin{abstract}
%%%
High-fidelity experimental characterization of turbulent premixed flames remains severely limited by the cost and complexity of advanced diagnostics, particularly under elevated pressures and intense turbulence where measurements of coupled flame morphology and burning dynamics are sparse. Here, we develop a hierarchical multi-fidelity neural network framework (MuFiNNs) to address this challenge by integrating sparse high-fidelity experimental data with structured low-fidelity representations that encode dominant physical trends. The framework combines hierarchical low-fidelity construction with nonlinear multi-fidelity correction to learn coupled geometric and reactive flame behavior while recovering discrepancies that simplified models alone cannot capture. The methodology is applied to expanding turbulent premixed flames to predict both three-dimensional flame wrinkling dynamics and turbulent mass burning velocity across varying fuels, pressures, and turbulence intensities. Using experimentally informed low-fidelity trend models together with sparse high-fidelity measurements, MuFiNNs accurately reconstructs observed flame behavior, enables reliable interpolation across unseen operating conditions, and demonstrates robust extrapolative capability beyond the training domain. Importantly, the framework remains effective even in regimes characterized by noisy, weakly structured, or experimentally inaccessible data, where conventional data-driven approaches often fail. These results show that hierarchical multi-fidelity learning provides a scalable and physically grounded strategy for predictive combustion modeling in data-limited regimes. More broadly, this work establishes multi-fidelity scientific machine learning as a practical framework for extracting physically meaningful predictive models from sparse experiments, particularly for instability-dominated and turbulence-sensitive reactive flows where conventional high-fidelity data acquisition is prohibitively demanding.
%%%%
\end{abstract}


\end{frontmatter}

%\linenumbers

\section*{Novelty and significance statement}
This work presents a hierarchical multi-fidelity learning framework for predicting coupled flame morphology and turbulent burning response in expanding premixed flames from sparse high-fidelity experimental data. The novelty lies in integrating experimentally informed low-fidelity trend models with nonlinear high-fidelity measurements to jointly predict 3D flame-surface evolution and turbulent mass burning velocity, rather than treating flame wrinkling and burning rate as separate observables. The framework is evaluated using methane and hydrogen flame datasets spanning fuel composition, equivalence ratio, pressure, temperature, and turbulence intensity, with systematic hold-out tests for interpolation, extrapolation, and unseen operating conditions. The results show that multi-fidelity learning can recover physically consistent flame behavior in data-limited regimes where direct 3D diagnostics and pressure-based burning-velocity measurements are sparse, noisy, or experimentally challenging. This establishes a practical route for predictive combustion modeling that combines sparse experiments with structured surrogate physics for instability-sensitive turbulent premixed flames.

%% main text
% ===================== Main text =====================
\section{Introduction}

The rapid transition toward hydrogen and low-carbon fuels has renewed interest in the dynamics of turbulent premixed flames, particularly in regimes where intrinsic flame-front instability coexists with externally imposed turbulence. Lean premixed hydrogen--air flames form a canonical low-Lewis-number system ($Le<1$), in which preferential diffusion, flame stretch sensitivity, and gas expansion jointly destabilize the flame front. Classical analyses link diffusive--thermal imbalance and the Markstein response to the amplification and wavelength selection of finite disturbances \cite{Zeldovich1944,Markstein1951,Clavin1985,MatalonMatkowsky1982,Law2006,BechtoldMatalon2001}, while hydrodynamic instability and its nonlinear evolution are captured by the classical framework and related weakly nonlinear theories \cite{Sivashinsky1977,MichelsonSivashinsky1977,BychkovLiberman2000}. As a consequence, low-$Le$ flames can develop cellular structures, curvature localization, and persistently non-Gaussian curvature statistics even in nominally laminar, outwardly propagating conditions \cite{WilliamsBook,Peters2000,PoinsotVeynante2005}.

A key practical implication is that instability-driven morphology does not merely 'decorate' a laminar flame; it can measurably modify the global consumption rate through surface amplification and curvature-conditioned transport. A sequence of recent studies has established (i) stretch- and pressure-controlled cellular onset in outwardly propagating hydrogen flames \cite{Xie2022FuelCellular,LiXie2024FuelEthaneH2}, (ii) self-acceleration and global pulsation driven purely by intrinsic surface amplification \cite{Xie2023FuelPulsation}, and (iii) wrinkling/self-disturbance dynamics with a pronounced convexity bias toward reactant-facing protrusions \cite{Xie2024CnFwrinkling}. Importantly, fully three-dimensional characterization shows that instability evolution in 3D differs fundamentally from 2D projections, with nonlinear mode competition, curvature redistribution, and topology bifurcation that only become evident with volumetric reconstructions \cite{Xie2025CnF3D}. These findings echo broader evidence that differential diffusion can imprint a directional morphology and alter local burning rates via curvature and strain \cite{Matalon2009,ChenIm2018}.

When turbulence is imposed, the interaction between turbulence-driven wrinkling and intrinsic instability becomes strongly nonlinear and condition-dependent. In classical turbulent premixed combustion, turbulent burning velocity is related to flame surface area growth and flame surface density (FSD), through flamelet concepts and surface-transport formulations \cite{Peters2000,PoinsotVeynante2005,VeynanteVervisch2002,CandelPoinsot1990,Bray1990}. These frameworks often assume that turbulence is the primary source of corrugation and that chemistry remains flamelet-like. However, in low-$Le$ hydrogen flames, intrinsic instability provides an 'open' amplification pathway: turbulence can act as a catalyst that injects energy into an already unstable morphological state, selectively enhancing convex leading structures and increasing the prevalence of cusp-like features, rather than producing a statistically symmetric wrinkling field \cite{Aspden2011,ChenIm2018}. In contrast, thermo-diffusively stable flames (e.g.\ methane under conditions yielding stabilizing diffusive response) typically exhibit turbulence-driven area increase without generating a persistent population of sharply peaked convex structures \cite{Bradley2003,HowarthAspden2012}. This conditional response implies that models based solely on area scaling or single-parameter wrinkling factors may be insufficient when intrinsic instability and turbulence operate on comparable time scales.

A central difficulty in the analysis of turbulent premixed flames is the identification of observables that jointly characterize flame morphology and global burning response. In the present work, the normalized three-dimensional flame surface area, $A_{3D}/a_{3D}$, and the turbulent mass burning velocity, $u_{tm}$, are selected because together they provide a coupled geometric--reactive description of the flame. Here, $A_{3D}$ is the reconstructed wrinkled flame surface area and $a_{3D}=4\pi r_{3D}^{2}$ is the surface area of an equivalent smooth sphere with the same burned-gas volume, so that $A_{3D}/a_{3D}$ directly measures the departure from a smooth reference geometry and hence the degree of flame wrinkling and surface amplification. This is particularly relevant in lean hydrogen flames, where flame-surface corrugation is governed not only by turbulence-induced folding but also by intrinsic thermo-diffusive and hydrodynamic instabilities \cite{Peters2000, PoinsotVeynante2005, VeynanteVervisch2002, Xie2025CnF3D, Ahmed2021}. By contrast, $u_{tm}$ characterizes the effective global burning response of the flame and reflects the integrated outcome of flame-surface growth, preferential diffusion, curvature effects, flame stretch, and instability-mediated transport \cite{CandelPoinsot1990,Bray1990,Ahmed2021}. The combined consideration of $A_{3D}/a_{3D}$ and $u_{tm}$ is therefore essential for assessing whether an increase in burning rate can be explained by flame-area amplification alone or whether additional mechanisms contribute to the enhancement of turbulent burning \cite{Ahmed2021}. From a measurement perspective, however, these quantities are also among the most demanding to obtain simultaneously. Accurate determination of $A_{3D}/a_{3D}$ requires time-resolved three-dimensional flame reconstruction using advanced multi-laser-sheet diagnostics and substantial post-processing, including volumetric reconstruction, triangulation, and surface analysis, with uncertainties that become particularly severe for highly wrinkled flames and under intense turbulence \cite{Ahmed2024PoF}. The extraction of $u_{tm}$ from pressure-rise measurements is likewise subject to non-negligible uncertainty because it depends on signal quality, temporal differentiation, smoothing, and the kinematic and volumetric assumptions adopted during post-processing; these issues are especially important at small flame radii and in unstable flames, where pressure-based and geometry-based characterizations may diverge \cite{Xie2025CnF3D,Ahmed2021,Zheng2025CnF}. Consequently, datasets containing simultaneous measurements of $A_{3D}/a_{3D}$ and $u_{tm}$ over broad ranges of pressure, temperature, equivalence ratio, and turbulence intensity are inherently sparse, expensive, and heterogeneous. This combination of strong physical relevance and severe data limitation directly motivates the use of machine-learning-based multi-fidelity frameworks that can preserve the coupling between flame geometry and global burning while making efficient use of limited high-fidelity measurements \cite{Xie2025CnF3D,Ahmed2021,Ahmed2024PoF,Zheng2025CnF}.

Recently, advances in computational power and data acquisition have enabled the widespread adoption of data-driven modeling approaches across science and engineering. In this context, artificial intelligence (AI) and machine learning (ML) have emerged as scalable and flexible tools for developing predictive models of complex physical systems. In particular, ML methods have shown a strong ability to capture highly nonlinear relationships where traditional approaches face limitations. Their ability to learn complex mappings directly from data makes them well suited for modeling multiscale and strongly coupled physical phenomena, enabling faster prediction and data-driven discovery across a wide range of applications \cite{Raissi2019PINN,Karniadakis2021PIML,Raissi2020HiddenFluid,Raissi2018DeepHiddenPhysics}.

Following these developments, a wide range of ML algorithms have been proposed, differing in their structure, input--output representation, and target applications. Among these, neural networks (NNs) have attracted significant attention due to their capacity to represent complex nonlinear relationships within data \cite{Zhang2016RandomizedNN}. Inspired by biological neural systems, NNs consist of interconnected processing units that learn functional mappings between inputs and outputs through adaptive optimization of network parameters. Various architectures, including ANNs, DNNs, CNNs, and RNNs, have demonstrated strong performance across diverse physical applications \cite{Mahmoudabadbozchelou2021MFNN}.

Despite these successes, standard data-driven approaches face several fundamental limitations in scientific settings. In particular, neural networks typically require large volumes of high-quality training data to achieve reliable predictive performance, which is often unavailable in practice. Moreover, such models are generally limited to interpolation within the range of the training data and may fail to generalize under extrapolative conditions. Another key limitation is the lack of embedded physical constraints, as conventional ML models primarily learn statistical correlations rather than governing principles, often leading to reduced interpretability and robustness \cite{Carvalho2019,Lennon2023,Buhrmester2021}.

To address these challenges, physics-informed machine learning frameworks have been developed to incorporate prior physical knowledge into the learning process. A prominent example is the Physics-Informed Neural Network (PINN) framework \cite{Raissi2019PINN,Karniadakis2021PIML}, where governing equations are embedded into the loss function as soft constraints. By enforcing physical consistency, PINNs reduce reliance on large datasets and improve generalization in data-scarce regimes. Recent developments have extended these approaches to more complex settings, including forward and inverse modeling without requiring labeled data for internal fields \cite{Zolfaghari2026NonlocalPINN, Mahmoudabadbozchelou2022nnPINNs}. Several variants have also been proposed, including PPINN for parallel-in-time learning \cite{Meng2019PPINN}, MPINN for multi-fidelity data integration \cite{Meng2020MPINN,Perdikaris2017MF,Forrester2007MF,Mahmoudabadbozchelou2021MFNN, Mahmoudabadbozchelou2022DigitalRheometerTwins, Saadat2024MFNN}, and fPINN for fractional differential equations \cite{Dabiri2023RhINN,Dabiri2025fPINNReview,Pang2019fPINN}, along with approaches such as nPINNs and DeepONet for modeling nonlocal and operator-based systems \cite{Lu2019DeepONet,Pang2020nPINN, Mangal2025}. In parallel, alternative paradigms such as neural operator learning have been introduced to directly learn solution operators of complex systems, enabling efficient surrogate modeling across varying conditions \cite{saberi2025rheoformer}. 

In general, physical knowledge can be incorporated into neural networks either explicitly, through governing differential equations, or implicitly, through structured model design and inductive biases. While explicit approaches are effective when accurate constitutive relations are available, implicit strategies are more suitable when the underlying physics is complex, partially known, or difficult to enforce directly.  In settings where data are limited and physics is only partially known, multi-fidelity learning provides an effective alternative by combining data from multiple sources with different levels of accuracy. In this framework, low-fidelity (Lo-Fi) data capture dominant trends and provide a structured baseline, while high-fidelity (Hi-Fi) data—though limited and expensive to obtain—serve to correct systematic discrepancies and anchor predictive accuracy \cite{Saadat2024MFNN, Mahmoudabadbozchelou2021MFNN}. Lo-Fi data can be generated from simplified models, empirical correlations, or synthetic approximations, whereas Hi-Fi data typically originate from high-resolution experiments or detailed simulations. This hierarchical integration enables physically consistent, robust, and data-efficient modeling, particularly in problems where neither purely data-driven nor fully physics-constrained approaches are sufficient.

In this work, we develop a data-driven framework to model the coupled geometric and reactive behavior of turbulent premixed hydrogen flames across a wide range of thermodynamic conditions. Leveraging limited high-fidelity measurements of the three-dimensional flame surface area $A_{3D}$, flame surface radius $r_{3D}$, and turbulent mass burning velocity $u_{tm}$, we employ a multi-fidelity learning strategy to overcome data scarcity and capture their nonlinear interdependence across operating regimes. This work is motivated by the lack of predictive models capable of consistently capturing the interaction between flame morphology and global burning rate in regimes where intrinsic instability and turbulence coexist, particularly in the absence of sufficient high-fidelity data and reliable physics-based closures. 

\section{Problem setup and Methodology}

Here, we present a physics-guided, data-driven modeling framework that leverages neural network (NN) capabilities to predict experimentally measured observables from limited high-quality data. To this end, we adopt a \textit{Multi-Fidelity Neural Network} (MuFiNNs) architecture that systematically integrates structured low-fidelity approximations with sparse high-fidelity experimental measurements. The objective is to construct a surrogate model that preserves the underlying physical structure of the system while achieving quantitative agreement with high-fidelity observations across a range of operating conditions. In the present study, the MuFiNNs framework is trained and evaluated using a limited set of turbulent premixed flame datasets for methane and hydrogen under varying thermodynamic and turbulence conditions. The experimental database spans a range of equivalence ratios, temperatures, pressures, and turbulence intensities, thereby enabling assessment of model performance across distinct combustion regimes. The complete operating conditions are summarized in Table~\ref{tab:cases}.

For each operating condition, the experimentally measured high-fidelity (Hi-Fi) quantities include the time-resolved evolution of the flame radius $r(t)$, the wrinkled flame surface area $A(t)$, and the turbulent burning velocity $u_{tm}$. Consequently, the dataset exhibits heterogeneous input–output mappings, involving both temporal evolution and parameter-dependent relationships, which necessitates a flexible learning architecture. These measurements provide physically accurate but sparsely sampled information across turbulence intensities and different stages of flame development. This heterogeneity, combined with the limited availability of high-fidelity data, poses a significant challenge for conventional modeling approaches and motivates the adoption of a multi-fidelity learning framework.

The low-fidelity (Lo-Fi) representations are synthetically generated using regression-based approximations designed to capture the dominant structural trends of the experimental data. Depending on the target variable, these Lo-Fi models are constructed using linear regression and second-order two-dimensional polynomial fits. These approximations provide smooth and physically interpretable trend representations that preserve large-scale behavior but do not fully reproduce the nonlinear complexities observed in the Hi-Fi measurements. The MuFiNNs architecture subsequently learns the corrective mapping between these regression-based Lo-Fi approximations and the experimentally measured Hi-Fi data across all investigated combustion regimes. By training across multiple fuels and thermodynamic conditions, the framework is designed to evaluate both interpolation within measured operating ranges and generalization across distinct combustion environments.

\subsection{Multi-Fidelity Framework}

In a multi-fidelity surrogate setting, the multi-fidelity prediction, denoted here as $y_{\text{MF}}$, can be interpreted as a combination of the low-fidelity prediction $y_{\text{LF}}$ and the high-fidelity input variables $x_{\text{HF}}$. A classical approach assumes that the high-fidelity response can be represented as a corrected version of the low-fidelity approximation. This idea is often expressed through a linear correlation structure,

\begin{equation}
y_{\text{HF}} = \rho(x_{\text{HF}})\, y_{\text{LF}} + \delta(x_{\text{HF}})
\label{eq:linear_MuFiNNs}
\end{equation}

where $\rho(\cdot)$ and $\delta(\cdot)$ denote multiplicative and additive correlation functions, respectively. In this formulation, $\rho$ and $\delta$ are typically calibrated to map the low-fidelity response to the high-fidelity level. These correlation functions are problem-dependent and do not possess predefined analytical forms. Depending on the application, they may be approximated using polynomial functions, basis expansions, or data-driven models whose coefficients are determined from the available data.

While the linear correction model provides a useful starting point, combustion systems often exhibit strongly nonlinear interactions between turbulence intensity, flame geometry, and thermodynamic state. As a result, the discrepancy between low- and high-fidelity representations cannot generally be captured through linear scaling alone. A more flexible representation is therefore introduced,

\begin{equation}
y_{\text{HF}} = \mathcal{N}(x_{\text{HF}}, y_{\text{LF}})
\label{eq:general_MuFiNNs}
\end{equation}

where $\mathcal{N}(\cdot)$ denotes a nonlinear transformation that maps the low-fidelity approximation and high-fidelity input variables to the high-fidelity response. In the proposed MuFiNNs architecture, this nonlinear operator is decomposed into two complementary components,

\begin{equation}
y_{\text{HF}} =
\mathcal{N}_{L}(x_{\text{HF}}, y_{\text{LF}})
+
\mathcal{N}_{NL}(x_{\text{HF}}, y_{\text{LF}})
\label{eq:decomposed_MuFiNNs}
\end{equation}

where $\mathcal{N}_{L}$ captures the primary linear correction between fidelity levels, while $\mathcal{N}_{NL}$ accounts for nonlinear discrepancies that cannot be represented through simple scaling. This decomposition improves training stability by separating global trend correction from higher-order effects, enabling the nonlinear branch to capture complex interactions associated with turbulence intensity, flame geometry, and thermodynamic state.

\begin{figure*}[t]
    \centering
    \includegraphics[trim=20 5 20 5, clip,width=0.7\linewidth]{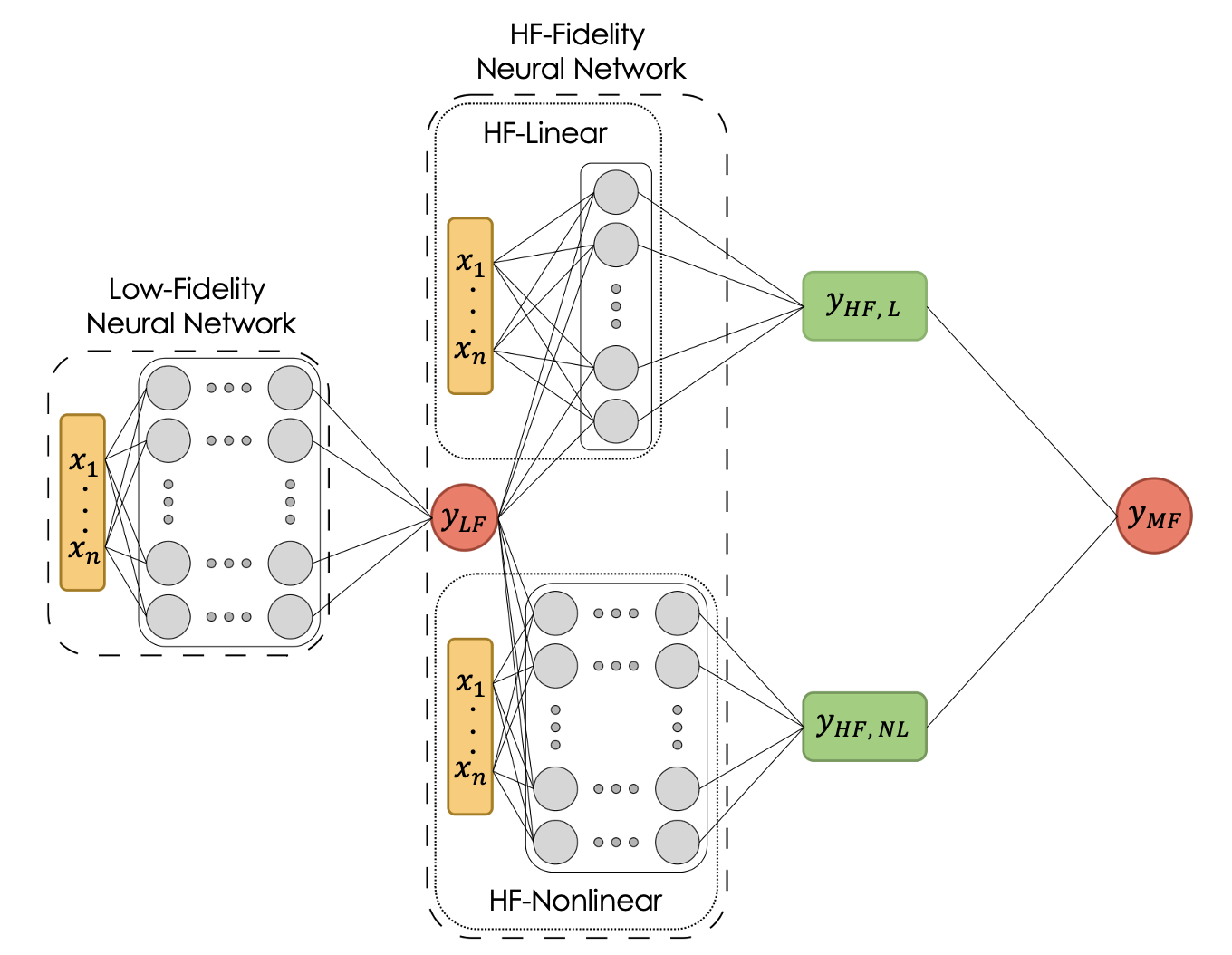}
    \caption{Schematic representation of the Multi-Fidelity Neural Network (MuFiNNs) architecture developed in the present work. The $n$ number of input variables $x_i$ are first processed through the low-fidelity (LF) neural network to synthetically generate a structured approximation, $y_{\text{LF}}$. This LF prediction, together with the original input variables, is then fed into two high-fidelity correction branches: a linear network and a nonlinear network. The final multi-fidelity prediction, $\hat{y}_{\text{MF}}$, is obtained by summing the outputs of these two branches.}    
    \label{fig:MuFiNNs_architecture}
\end{figure*}

The schematic representation of the proposed MuFiNNs architecture is shown in Fig.~\ref{fig:MuFiNNs_architecture}. The framework consists of three neural networks: a low-fidelity (LF) network and two high-fidelity (HF) correction branches, namely a linear branch and a nonlinear branch. This general structure defines the multi-fidelity correction mechanism adopted throughout the present study. In the present work, multiple MuFiNNs models are constructed to address different prediction tasks and operating conditions. While the set of input variables and the overall architectural structure remain consistent with that illustrated in Fig.~\ref{fig:MuFiNNs_architecture}, the number of hidden layers and neurons per layer are selected individually for each model. This task-specific configuration reflects the varying complexity of the underlying input--output relationships associated with each prediction problem. The architectural hyperparameters are therefore tailored to the complexity of the respective task, while preserving a unified multi-fidelity correction framework across all cases.

Accordingly, the low-fidelity neural network, $\mathcal{N}_{\mathrm{LF}}$, takes the LF input variables and predicts a structured low-fidelity response, $y_{\mathrm{LF}}$, which captures the dominant global behavior of the turbulent burning parameters across the input space. The LF output $y_{\mathrm{LF}}$, together with the original input vector $\mathbf{x}$, is then supplied to the linear and nonlinear HF correction branches. The linear branch produces $y_{\mathrm{HF},L}$ and captures first-order correlations between fidelity levels, while the nonlinear branch produces $y_{\mathrm{HF},NL}$ and models higher-order interactions and localized discrepancies.

The final multi-fidelity prediction is obtained as
\begin{equation}
y_{\mathrm{MF}} = y_{\mathrm{HF},L} + y_{\mathrm{HF},NL}
\label{eq:final_prediction}
\end{equation}

where $y_{\mathrm{HF},L}$ and $y_{\mathrm{HF},NL}$ denote the outputs of the linear and nonlinear correction networks, respectively. The distinction between fidelity levels is rooted in the data construction strategy. The high-fidelity (Hi-Fi) data are predetermined by the available experimental measurements for each flow condition. In contrast, the construction of the low-fidelity (Lo-Fi) dataset is user-defined and depends on the chosen simplified representation of the physical system. In the present study, the Lo-Fi data are synthetically generated from global trend models designed to capture the dominant structural behavior of the turbulent burning parameters. The methodology used for Lo-Fi data generation is described in the next section, along with further implementation details. The trainable parameters of the MuFiNNs architecture are obtained by minimizing a compound loss function defined as

\begin{equation}
\phi =
\text{MSE}_{LF}
+
\text{MSE}_{HF}
+
\lambda_{LF} \sum w_{LF}^2
+
\lambda_{HF,NL} \sum w_{HF,NL}^2
\label{eq:loss_function}
\end{equation}

where the mean squared error (MSE) is defined as

\begin{equation}
\text{MSE} = \frac{1}{N} \sum_{i=1}^{N} \left( y_{\text{actual},i} - y_{\text{predicted},i} \right)^2
\label{eq:mse_definition}
\end{equation}

and represents the deviation between predicted and actual data. In Eq.~(\ref{eq:loss_function}), $\text{MSE}_{LF}$ and $\text{MSE}_{HF}$ denote the deviations between predicted and measured data for the low- and high-fidelity datasets, respectively. In Eq.~(\ref{eq:loss_function}), the first two terms enforce consistency with both fidelity levels, ensuring that the model remains faithful to the structured trends embedded in the Lo-Fi data while simultaneously matching the experimentally measured Hi-Fi observations. The last two terms correspond to $L_2$ regularization (weight decay) applied to the LF and nonlinear high-fidelity branches, respectively, where $\lambda_{LF}$ and $\lambda_{HF,NL}$ are regularization coefficients. These regularization terms penalize large weight magnitudes, thereby preventing overfitting and improving generalization, particularly when the high-fidelity dataset is limited.

In this framework, training of the MuFiNNs models is performed using a hybrid optimization strategy. In the first stage, all trainable parameters are optimized using the Adam optimizer with a warmup--cosine learning-rate schedule and gradient clipping to ensure stable convergence of the coupled low- and high-fidelity loss terms. To prevent unnecessary over-training and improve computational efficiency, a plateau-based stopping criterion is applied during the Adam stage. For each case, training is terminated when the variation of the total loss over a moving window falls below a prescribed threshold, resulting in a case-dependent number of iterations determined automatically by the loss behavior. In the second stage, the solution is refined using the L-BFGS algorithm implemented through TensorFlow Probability. This quasi-Newton method leverages curvature information to accelerate local convergence and improve minimization of the multi-fidelity loss function. All trainable parameters of the LF network and both the linear and nonlinear HF correction branches are optimized jointly during both stages. This two-step optimization strategy enhances numerical stability and improves final predictive accuracy.

The multi-fidelity formulation enables the model to benefit simultaneously from the accuracy of the high-fidelity data and the abundance of the low-fidelity data. Because experimentally measured Hi-Fi samples are comparatively sparse, the LF network provides global structural trends across the input space, guiding the HF correction branches during training. This guidance stabilizes the optimization process and reduces the risk of divergence toward non-physical solutions. As a result, the MuFiNNs architecture achieves a balanced trade-off between physical structure preservation, experimental accuracy, and predictive generalization.

\subsection{Training Datasets}

The high-fidelity (Hi-Fi) dataset consists of three complementary data components: (i) geometric information derived from the reconstructed three-dimensional flame surface, (ii) volume-based measures of flame growth, and (iii) pressure-based turbulent burning-velocity measurements.  The geometric component is obtained from imaging-based three-dimensional flame reconstructions. The wrinkled flame surface area $A_{3D}(t)$ is computed directly from the triangulated 3D flame surface. For each case, the reconstructed flame also provides the time-resolved burnt volume $V_{3D}(t)$, from which a volume-equivalent radius $r_{3D}(t)$ is obtained. These quantities characterize the geometric evolution of the expanding turbulent flame. In particular, $r_{3D}$ is used to define the smooth reference area $a_{3D}(t)=4\pi r_{3D}^2$, enabling evaluation of the wrinkling ratio $A_{3D}/a_{3D}$ as mentioned earlier.
The third component of the Hi-Fi dataset is the turbulent burning velocity $u_{tm}$, which is computed from the measured pressure trace using a thermodynamic (fractional burning-rate) formulation based on $\mathrm{d}P/\mathrm{d}t$, yielding a time-resolved signal $u_{tm}(t)$. However, since the flame radius provides a geometry-based coordinate that is directly linked to flame surface area and burning dynamics---unlike time, which depends on the specific evolution rate of each experiment---the burning velocity is re-parameterized using the corresponding radius history to obtain $u_{tm}(r)$. This transformation removes dependence on the experimental time scale and enables a geometry-consistent representation of the burning process, allowing meaningful comparison across different operating conditions.

In contrast, the low-fidelity (Lo-Fi) dataset is synthetically generated using a simplified physics-inspired trend model that approximates the dominant dependence of the system on the governing variables, namely flame radius and turbulence intensity. The Lo-Fi representation captures the primary structural trends observed in the experiments while remaining computationally inexpensive and densely sampled over the domain.

To rigorously evaluate generalization, the Hi-Fi data corresponding to selected turbulence intensity levels are excluded during training. These masked cases are reserved exclusively for testing, enabling assessment of both interpolation within the observed range and extrapolation beyond the training regime. This strategy provides a comprehensive evaluation of the predictive capability of the MuFiNNs framework.

The experimental database considered in this study consists of seven operating conditions, summarized in Table~\ref{tab:cases}, including methane (CH$_4$) and hydrogen (H$_2$) mixtures spanning different equivalence ratios $\phi$, initial temperatures $T$, pressures $P$, and turbulence intensity ranges $u'$. These cases provide a broad range of thermo-chemical and flow conditions for assessing the robustness and generalization capability of the MuFiNNs framework.

\begin{table}[ht]
\centering
\caption{Summary of the investigated combustion cases and operating conditions.}
\label{tab:cases}
\begin{tabular}{llllll}
\toprule
Fuel & Case & $\phi$ (-) & $T$ (K) & $P$ (MPa) & $u'$ (m/s) \\
\midrule
CH$_4$ & I   & 0.60 & 365 & 0.1 & 0.3--1.5 \\
CH$_4$ & II  & 0.70 & 300 & 0.1 & 0.3--1.5 \\
CH$_4$ & III & 1.30 & 300 & 0.1 & 0.3--2.0 \\
CH$_4$ & IV  & 1.25 & 365 & 0.5 & 0.3--2.0 \\
H$_2$  & V   & 0.30 & 365 & 0.5 & 0.3--2.0 \\
H$_2$  & VI  & 0.40 & 365 & 0.5 & 0.3--1.5 \\
H$_2$  & VII & 0.30 & 360 & 0.1--1.0 & 0 \\
\bottomrule
\end{tabular}
\end{table}

\section{Results and Discussion}

In this section, the procedure for generating the Lo-Fi data for the target variables considered in this study is described for each case. Model performance is assessed through both interpolation and extrapolation tests, including the systematic masking of selected high-fidelity data during training. The predictive capability of the MuFiNNs framework is evaluated by comparing reconstructed and held-out cases, demonstrating its ability to generalize across different turbulence intensity levels.

\subsection{Flame Geometry: Surface Area and Radius ($A_{3D}$, $r_{3D}$)}

The geometric evolution of the turbulent flame is first examined through the three-dimensional surface area $A_{3D}$ and the corresponding volume-equivalent radius $r_{3D}$. These quantities provide a direct measure of flame growth and surface wrinkling, capturing the key geometric characteristics of the expanding turbulent flame.

For the methane (CH$_4$) cases (i--iv), the available data at each turbulence intensity level are limited, and as summarized in Table~\ref{tab:cases}, these cases correspond to different combinations of temperature, pressure, and equivalence ratio. Due to this limited data availability, directly training a single model across all conditions would not reliably capture the correct trends in the evolution of the geometric quantities. In particular, the model would struggle to disentangle the individual effects of the thermodynamic parameters on the flame behavior, limiting its ability to perform reliable interpolation and extrapolation across different operating conditions, which is the primary objective of this work.

\begin{figure*}[t]
\centering
\includegraphics[width=0.8\textwidth]{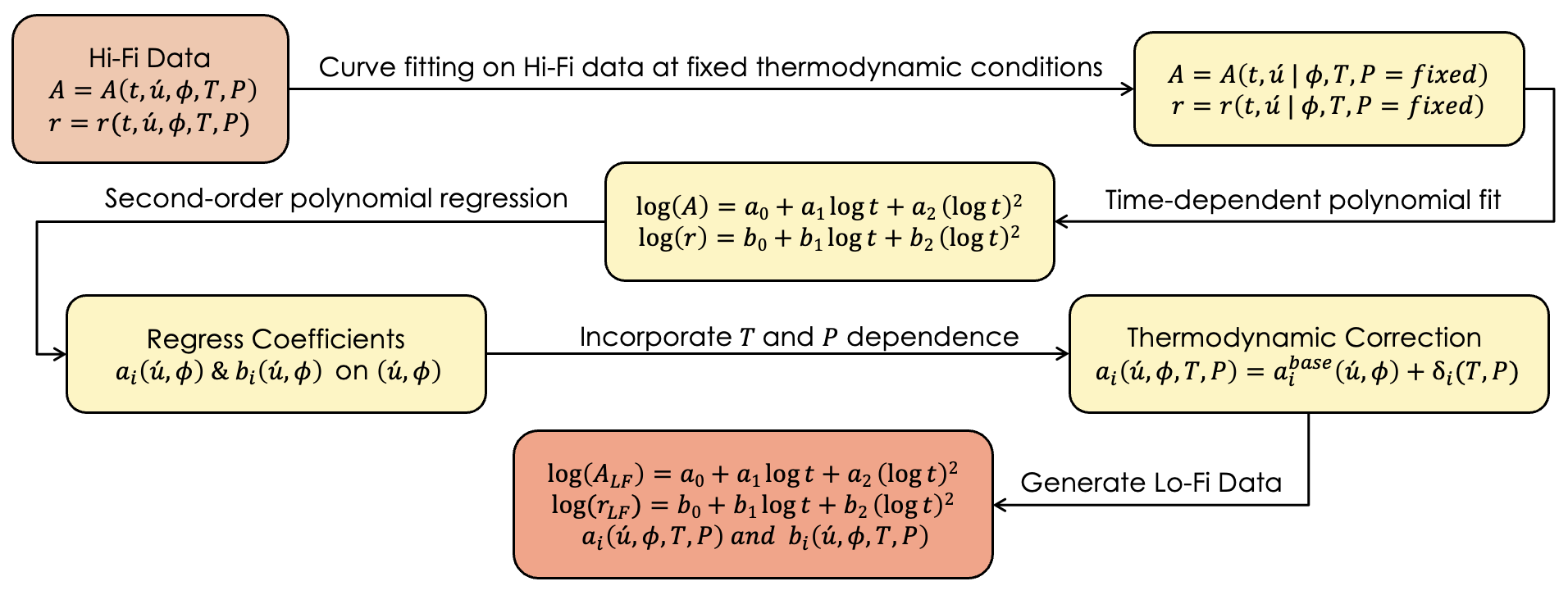}
\caption{
Schematic of the hierarchical construction of the low-fidelity (Lo-Fi) representation for $A_{3D}$ and $r_{3D}$, illustrating the stepwise regression of temporal evolution and subsequent parametric dependence on operating conditions.}
\label{fig:hierarchical_mfnn}
\end{figure*}

To address this limitation, a hierarchical multi-fidelity modeling strategy is adopted. In this approach, the system is trained in a structured, step-by-step manner, allowing it to progressively learn the influence of each parameter separately. This hierarchical structure reduces the complexity of the learning task by decoupling the effects of different parameters, enabling more stable and physically consistent regression under limited data conditions. A schematic representation of this procedure is shown in Fig.~\ref{fig:hierarchical_mfnn}.

Within this framework, the Hi-Fi data for $A_{3D}$ and $r_{3D}$ are first expressed as functions of time and operating conditions, i.e.,

\begin{equation}
A_{3D} = A_{3D}(t,u',\phi,T,P), \quad
r_{3D} = r_{3D}(t,u',\phi,T,P).
\end{equation}

Accordingly, the input space of the model consists of $(t,u',\phi,T,P)$, while the outputs correspond to the geometric quantities $A_{3D}$ and $r_{3D}$.

At fixed thermodynamic conditions $(\phi, T, P)$, curve fitting is performed to capture the temporal evolution of the flame geometry for each case separately. Specifically, the evolution of $A_{3D}$ and $r_{3D}$ is well approximated by a second-order polynomial in logarithmic time, and a second-order polynomial regression is therefore applied in logarithmic space, which provides a compact representation of the observed nonlinear growth trends over multiple time scales, yielding

\begin{equation}
\log(A_{3D}) = a_0 + a_1 \log t + a_2 (\log t)^2,
\label{eq:A_log_fit}
\end{equation}
\begin{equation}
\log(r_{3D}) = b_0 + b_1 \log t + b_2 (\log t)^2.
\label{eq:r_log_fit}
\end{equation}

The resulting coefficients $a_i$ and $b_i$ are then regressed as functions of turbulence intensity and equivalence ratio, i.e., $(u', \phi)$, to capture their parametric dependence within each thermodynamic case.

To account for variations in thermodynamic conditions, a correction term is introduced such that the coefficients are expressed as

\begin{equation}
a_i(u',\phi,T,P)
=
a_i^{\mathrm{base}}(u',\phi)
+
\delta_i(T,P),
\label{eq:a_correction}
\end{equation}
\begin{equation}
b_i(u',\phi,T,P)
=
b_i^{\mathrm{base}}(u',\phi)
+
\delta_i^{(r)}(T,P),
\label{eq:b_correction}
\end{equation}

where $a_i^{\mathrm{base}}(u',\phi)$ and $b_i^{\mathrm{base}}(u',\phi)$ denote the coefficients obtained at a reference thermodynamic condition, and $\delta_i(T,P)$ represents a thermodynamic correction term that captures the deviation of each case from this base trend. This formulation enables the model to systematically account for differences between cases with varying temperature and pressure, thereby improving consistency across the dataset.

This hierarchical construction enables the generation of a continuous and computationally efficient Lo-Fi representation of $A_{3D}$ and $r_{3D}$ across the parameter space, which is subsequently used within the MuFiNNs framework.

\begin{figure*}[t]
\centering
\includegraphics[trim=5 5 5 5, clip,width=\linewidth]{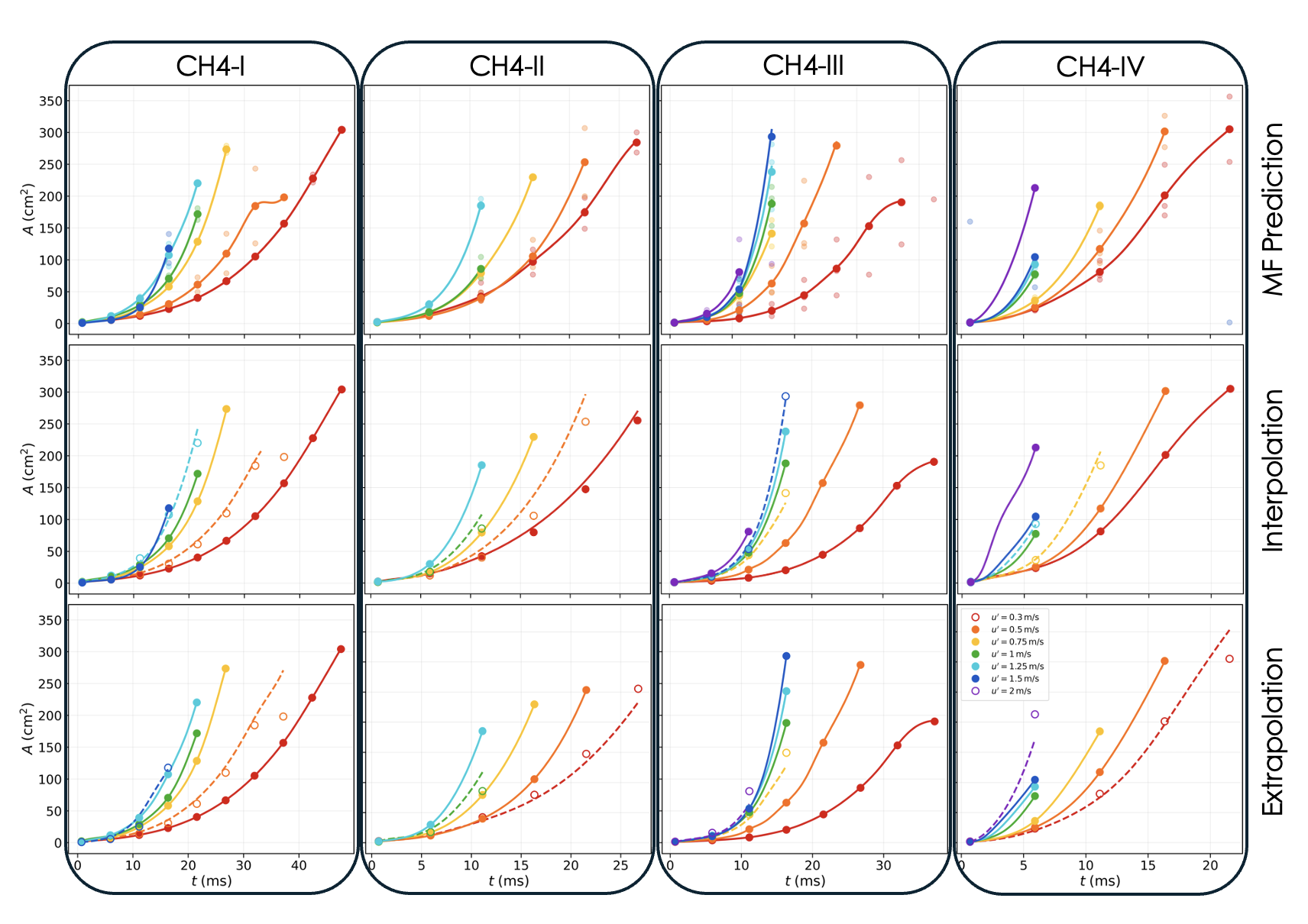}
\caption{MuFiNNs predictions of the flame surface area $A_{3D}(t)$ for methane (CH$_4$) cases (I--IV) across different turbulence intensities $u'$. 
Each column corresponds to a distinct operating condition, while the rows represent training (top), interpolation (middle), and extrapolation (bottom) assessments. 
Light-colored markers denote individual experimental realizations, and solid markers indicate their corresponding mean values used as high-fidelity (Hi-Fi) training targets. Solid lines represent MuFiNNs predictions. 
In the interpolation and extrapolation rows, filled markers indicate training data and hollow markers denote held-out data, while dashed curves correspond to model predictions for unseen turbulence intensity levels.}
\label{fig:CH4_A_surface_all}
\end{figure*}

Figure~\ref{fig:CH4_A_surface_all} presents the MuFiNNs predictions for the geometric evolution of the turbulent flame, characterized by the surface area $A_{3D}(t)$, across the methane (CH$_4$) cases (i--iv). The results are organized into three rows corresponding to training (top), interpolation (middle), and extrapolation (bottom) assessments.

In the top row, the MuFiNNs model is trained using the available high-fidelity (Hi-Fi) data. For each operating condition, three independent experimental measurements are available, shown as light-colored markers. The corresponding mean values, indicated by solid markers, are used as the Hi-Fi targets during training. As observed, the MuFiNNs predictions closely match the training data across all turbulence intensity levels, accurately capturing both the nonlinear temporal evolution and the dependence on turbulence intensity. The agreement between the predicted curves and the Hi-Fi data demonstrates that the model successfully learns the underlying geometric growth behavior of the flame despite the limited number of training samples.

To further evaluate the generalization capability of the model, an interpolation study is performed, as shown in the middle row of Fig.~\ref{fig:CH4_A_surface_all}. In this step, selected intermediate turbulence intensity levels are excluded from the training process. Specifically, for each case, two intermediate $u'$ conditions are removed entirely, and the model is trained only on the remaining data. The dashed curves represent the MuFiNNs predictions for these unseen conditions. Despite the absence of these data during training, the model is able to reconstruct the corresponding flame evolution with high accuracy. The predicted trends remain consistent with the experimental measurements, indicating that the MuFiNNs framework successfully captures the continuous dependence of flame geometry on turbulence intensity and can reliably interpolate within the observed parameter space. However, in certain cases (e.g., CH$_4$-I at $u'=0.5$ m/s), where the experimental data exhibit slight deviations from the overall trend of the dataset, the MuFiNNs predictions may not fully reproduce these local inconsistencies. This behavior is expected, as the model is trained to learn the dominant underlying trend rather than isolated experimental fluctuations. Nevertheless, the overall agreement remains strong, and the predicted evolution continues to follow the physically consistent trend of the system.

In the bottom row, the extrapolation capability of the model is examined. In particular, the model is tasked with predicting flame evolution at higher turbulence intensity levels that lie outside the range of the training data. This regime is of particular interest, as high-$u'$ conditions are experimentally challenging due to increased flame instability, measurement uncertainty, and limitations in optical diagnostics, as also reported in previous experimental studies. Despite these challenges, the MuFiNNs predictions remain physically consistent and follow the expected growth trends. Although minor deviations from the experimental data are observed at the highest turbulence levels, the overall agreement remains strong, demonstrating that the model can generalize beyond the training domain while preserving the correct qualitative and quantitative behavior.

\begin{figure}[t]
\centering
\includegraphics[width=0.9\linewidth]{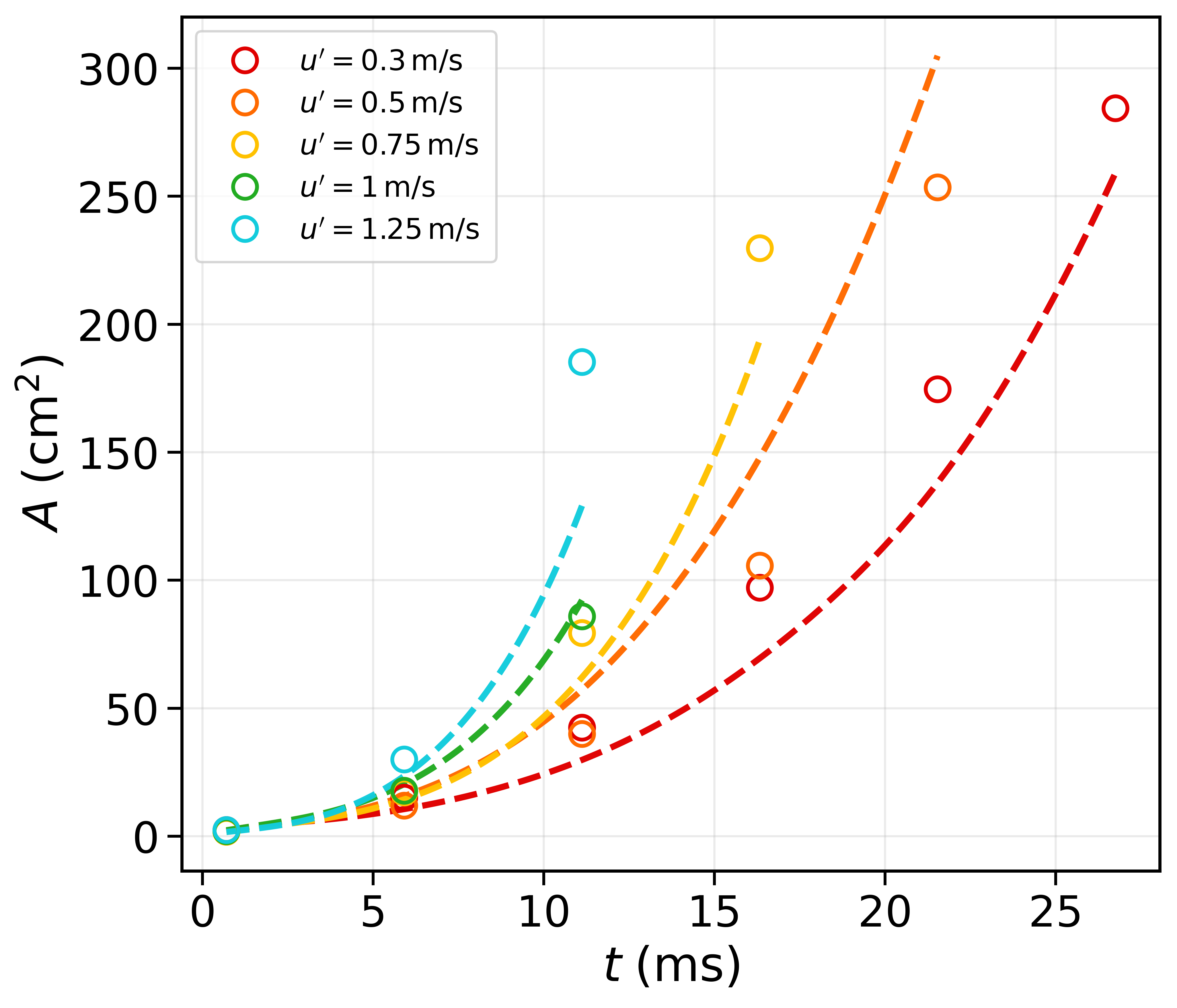}
\caption{MuFiNNs prediction of the flame surface area $A_{3D}(t)$ for the excluded methane case CH$_4$-II across different turbulence intensities $u'$. 
Symbols denote high-fidelity experimental measurements, while dashed lines represent model predictions obtained without using any data from this operating condition during training. }
\label{fig:predicted_case}
\end{figure}

To further assess the ability of the MuFiNNs framework to generalize across unseen thermodynamic conditions, an additional test is performed by entirely excluding one operating condition (CH$_4$-II) from the training dataset. This case is selected deliberately, as it shares the same pressure level with CH$_4$-I and a similar temperature range with CH$_4$-III, allowing the model to be exposed to partial thermodynamic information without directly observing this specific combination.

The model is then trained using the remaining cases and tasked with predicting the flame surface evolution for the excluded condition. The corresponding results are shown in Fig.~\ref{fig:predicted_case}. As observed, although the predictions do not exactly match the experimental measurements, the model is able to capture the correct qualitative behavior, including the growth trend and the overall magnitude of $A_{3D}(t)$.

This result highlights both the capability and the limitation of the current framework. On one hand, the MuFiNNs model successfully learns a structured representation of the flame dynamics, enabling it to generalize across partially unseen thermodynamic combinations. On the other hand, due to the limited number of available operating conditions in the dataset—specifically, only two distinct pressure and temperature levels—the model is constrained in its ability to fully reconstruct the precise quantitative response for completely unseen thermodynamic states.

Consequently, while the model can reliably capture the dominant trends and provide reasonable estimates within the explored parameter space, accurate prediction outside the sampled thermodynamic range requires a denser coverage of temperature and pressure conditions. This observation underscores the importance of incorporating a broader range of operating conditions in future datasets to further enhance the predictive capability of the multi-fidelity framework.

A similar analysis is performed for the hydrogen (H$_2$) cases (V and VI), yielding comparable predictive performance. However, since these cases vary only in equivalence ratio while sharing the same temperature and pressure, the learning problem is significantly simpler and does not require a hierarchical approach. 

Accordingly, the MuFiNNs model is trained only with respect to turbulence intensity $u'$, corresponding to the first stage of the hierarchical framework, and achieves accurate predictions across all conditions. Therefore, the methane cases—featuring variations in equivalence ratio, temperature, and pressure—are adopted as the primary benchmark, providing a more comprehensive and challenging test of the proposed framework.

For the final hydrogen case, corresponding to quiescent conditions ($u'=0$) with varying pressure, a simplified low-fidelity construction is adopted. Unlike the methane cases, where the flame evolution depends on multiple varying parameters and requires a hierarchical treatment, the present case involves only pressure variation under otherwise fixed conditions. As a result, the Lo-Fi representation is generated using a simple linear regression of the flame radius with time at each available pressure level, i.e., $r(t)\approx a(P)\,t+b(P)$. The pressure dependence of the linear coefficients is then inferred from the available training pressures and used to construct synthetic low-fidelity trends for unseen pressure conditions. This provides a computationally inexpensive baseline representation, upon which the multi-fidelity correction is learned.

\begin{figure*}[t]
\centering
\begin{minipage}{0.32\textwidth}
\begin{overpic}[width=\linewidth]{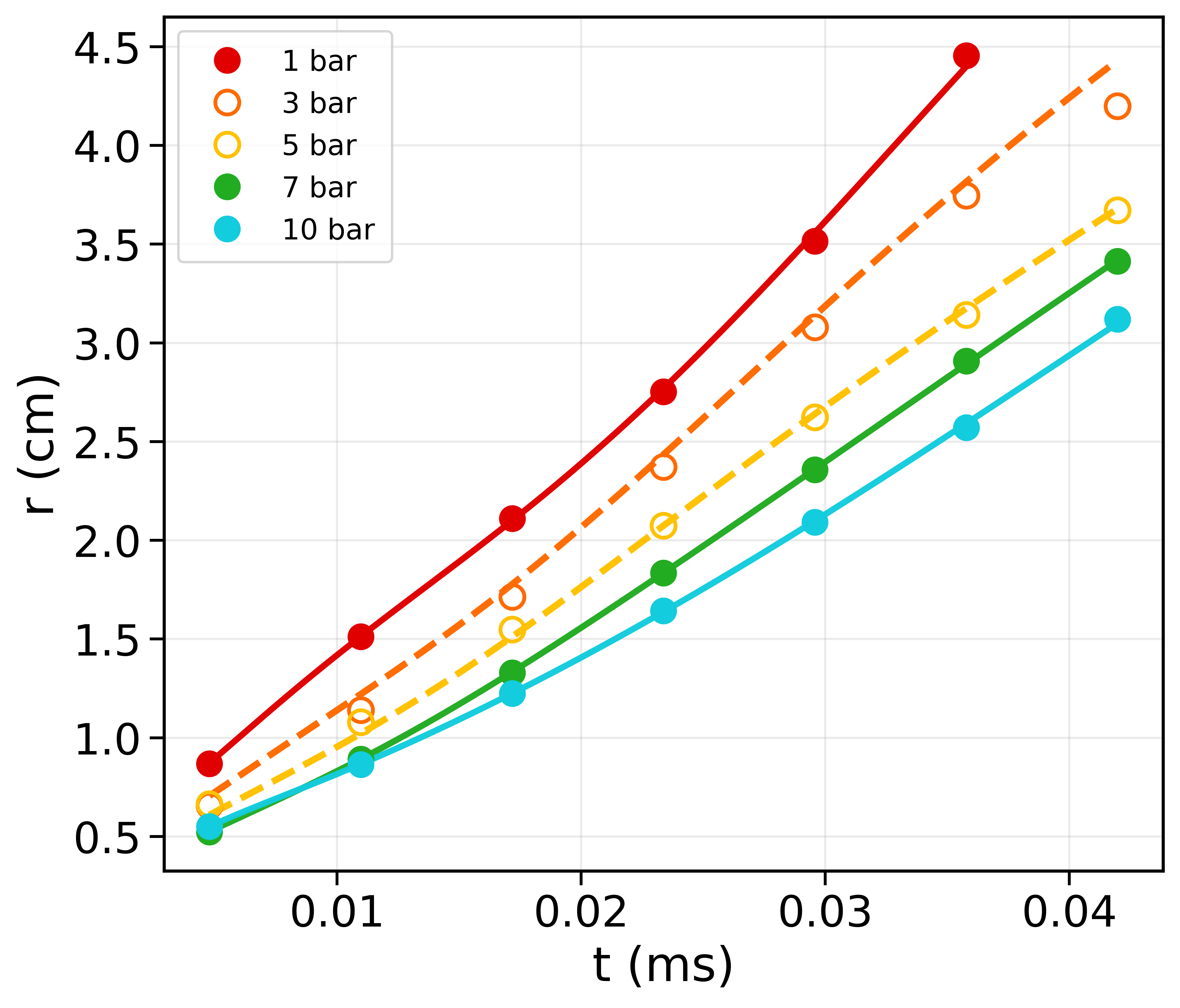}
\put(1,85){\textbf{(a)}}
\end{overpic}
\end{minipage}
\hfill
\begin{minipage}{0.32\textwidth}
\begin{overpic}[width=\linewidth]{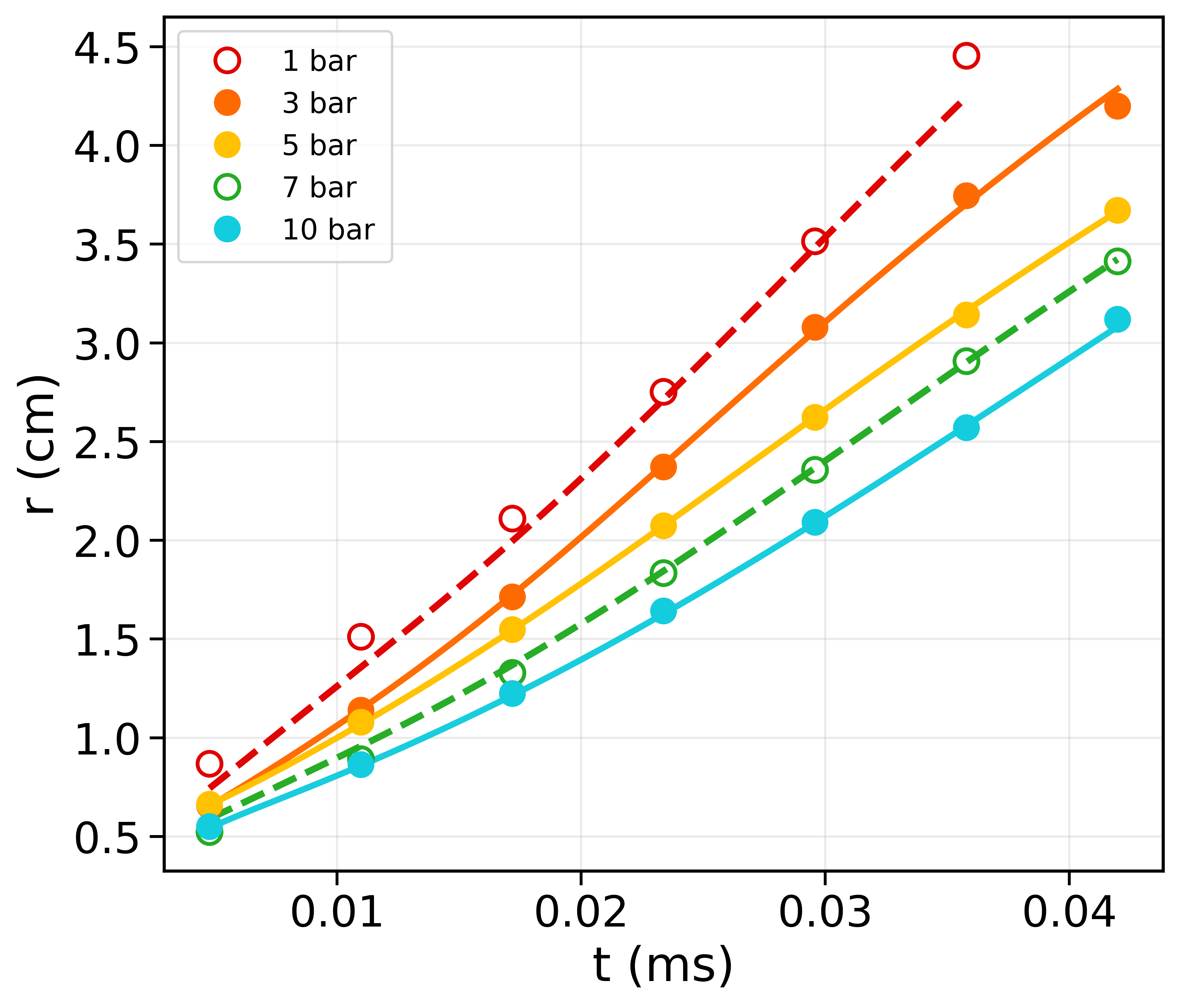}
\put(1,85){\textbf{(b)}}
\end{overpic}
\end{minipage}
\hfill
\begin{minipage}{0.32\textwidth}
\begin{overpic}[width=\linewidth]{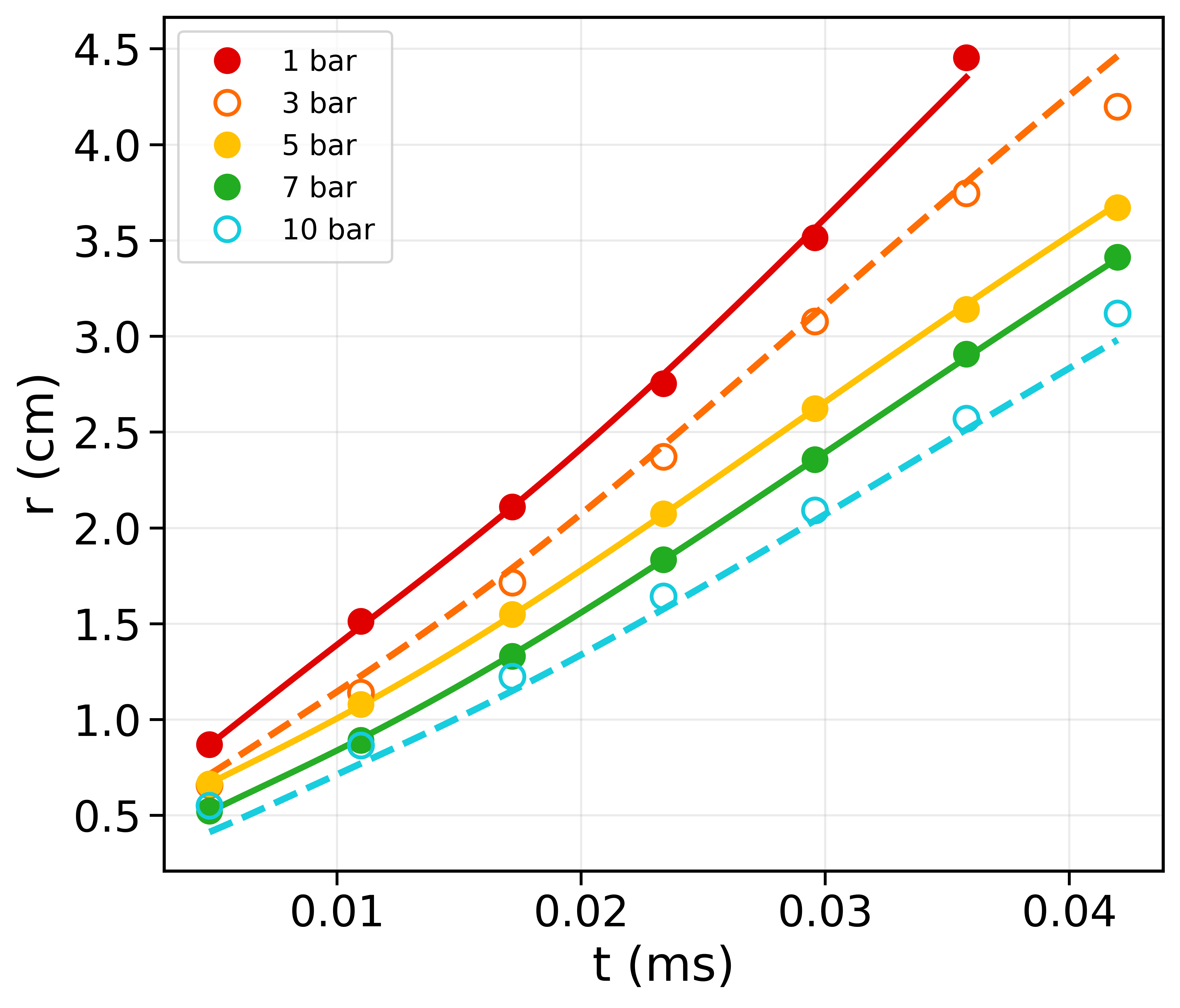}
\put(1,85){\textbf{(c)}}
\end{overpic}
\end{minipage}
\caption{Multi-fidelity predictions of the flame radius evolution $r(t)$ under varying chamber pressure, demonstrating the interpolation and extrapolation capability of the model. Filled markers indicate training data, while hollow markers denote held-out data. Solid curves correspond to MFNN predictions for pressures included in training, and dashed curves represent predictions for unseen (held-out) pressure conditions.
(a) Interpolation test, where intermediate pressures (3 and 5 bar) are excluded from training. (b,c) Combined interpolation–extrapolation tests, where selected pressure levels are removed from training, requiring prediction both within and beyond the range of the available data.}
\label{fig:radius_tests}
\end{figure*}

To assess the predictive capability of the framework under unseen pressure conditions, three hold-out studies are performed, as illustrated in Fig.~\ref{fig:radius_tests}. In the first case [Fig.~\ref{fig:radius_tests}(a)], two intermediate pressure levels are excluded from training in order to evaluate interpolation within the available pressure range. In the second case [Fig.~\ref{fig:radius_tests}(b)], the lowest pressure condition (1 bar) is excluded together with one additional intermediate pressure level (7 bar), allowing simultaneous assessment of interpolation and extrapolation toward the lower-pressure boundary of the domain. In the third case [Fig.~\ref{fig:radius_tests}(c)], the 3 bar and 10 bar conditions are removed, such that the model is required to extrapolate toward the highest pressure level. This final test is of particular importance, as generating reliable experimental data at elevated pressures is considerably more challenging in practice, and the ability to accurately predict flame evolution in this regime is therefore of significant value.

As shown by the results, the proposed framework accurately captures the pressure-dependent trends in flame-radius evolution across all scenarios. Both interpolation and extrapolation cases show excellent agreement with the experimental data. In particular, the predictions at high pressure remain physically consistent and reasonably accurate despite the absence of training data in this regime. These findings demonstrate that even with a simple regression-based Lo-Fi model, the MuFiNNs framework can reliably predict system behavior under experimentally challenging conditions.

\subsection{Turbulent Burning Velocity: $u_{tm}$}

The turbulent burning velocity $u_{tm}$ is examined as a key quantity describing the propagation of the turbulent flame. To enable consistent comparison across different operating conditions, the burning velocity is analyzed as a function of the flame radius, i.e., $u_{tm}(r)$, for all cases.

For the four cases CH$_4$-II, CH$_4$-III, H$_2$-V, and H$_2$-VI, the high-fidelity (Hi-Fi) turbulent mass burning velocity $u_{tm}$ is computed directly from the measured pressure traces using a thermodynamic fractional burning-rate formulation. Prior to this calculation, the raw pressure signals are denoised using a two-stage Savitzky--Golay filter and subsequently downsampled to improve the numerical stability of the time derivative $dP/dt$. To avoid post-peak oscillations and non-physical artifacts, the signals are truncated at the peak pressure $P_f$, retaining only the monotonic rising branch.

In the implemented form,

\begin{equation}
u_{tm}
=
\left(\frac{P_0}{P}\right)^{\frac{1}{\gamma_u}}
\left\{
1
-
\left(\frac{P_0}{P}\right)^{\frac{1}{\gamma_u}}
\left[
\frac{P_f - P}{P_f - P_0}
\right]
\right\}^{-\frac{2}{3}}
\;
\frac{R_0}{3(P_f - P_0)}
\frac{dP}{dt},
\end{equation}

where $P_0$ and $P_f$ denote the initial and peak pressures, respectively, $\gamma_u$ is the specific heat ratio of the unburned gas, and $R_0$ is the combustion chamber radius. The same formulation provides the corresponding flame radius as

\begin{equation}
r_m
=
R_0
\left\{
1
-
\left(\frac{P_0}{P}\right)^{\frac{1}{\gamma_u}}
\left[
\frac{P_f - P}{P_f - P_0}
\right]
\right\}^{\frac{1}{3}}.
\end{equation}

Thus, both $u_{tm}$ and $r$ are obtained consistently within the same pressure-based thermodynamic framework. The resulting $(r,u_{tm})$ pairs are sorted with respect to $r$ and restricted to the analysis window used for training. A second Savitzky--Golay smoothing is then applied to obtain the final Hi-Fi target curves $u_{tm}(r)$. While both the merged raw measurements and the smoothed signals are retained for diagnostic purposes, only the smoothed curves are used as training targets in the MuFiNNs framework.

The low-fidelity (Lo-Fi) dataset is synthetically generated using a structured global trend model inferred from the Hi-Fi training subset via linear regression. At each turbulence intensity level $u'$, the dependence of $u_{tm}$ on the flame radius $r$ is approximated by a linear relation, $u_{tm} \approx \alpha(u')\,r + \beta(u')$. The resulting coefficients are then regressed as smooth functions of $u'$ to capture the global dependence on turbulence intensity.

Using these parametric representations, a dense Lo-Fi dataset is constructed over a structured $(r,u')$ grid, yielding a smooth and computationally inexpensive approximation of the dominant trends in $u_{tm}(r,u')$ across the parameter space.

\begin{figure*}[t]
\centering
\begin{minipage}{0.32\textwidth}
\begin{overpic}[width=\linewidth]{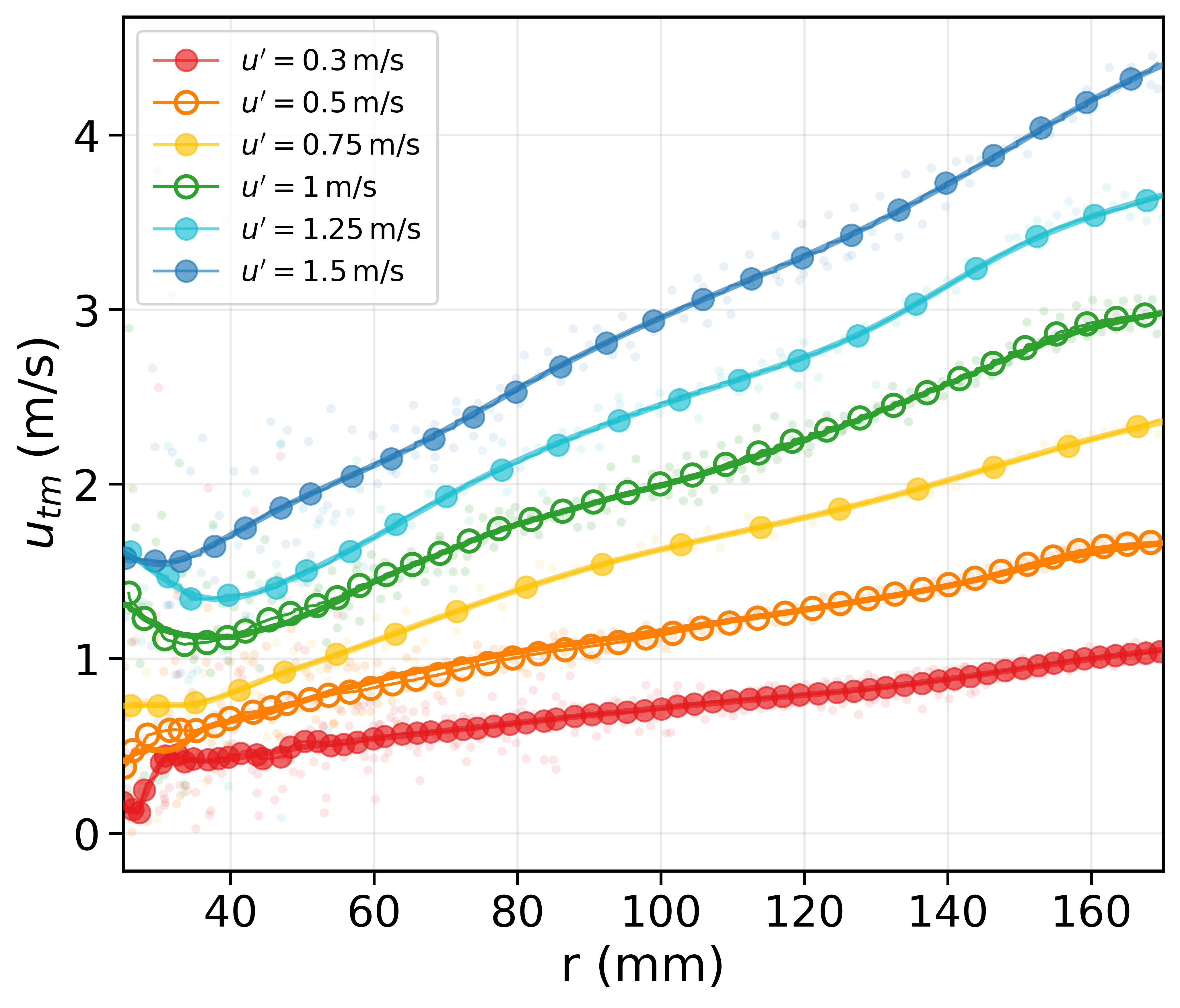}
\put(1,85){\textbf{(a)}}
\end{overpic}
\end{minipage}
\hfill
\begin{minipage}{0.32\textwidth}
\begin{overpic}[width=\linewidth]{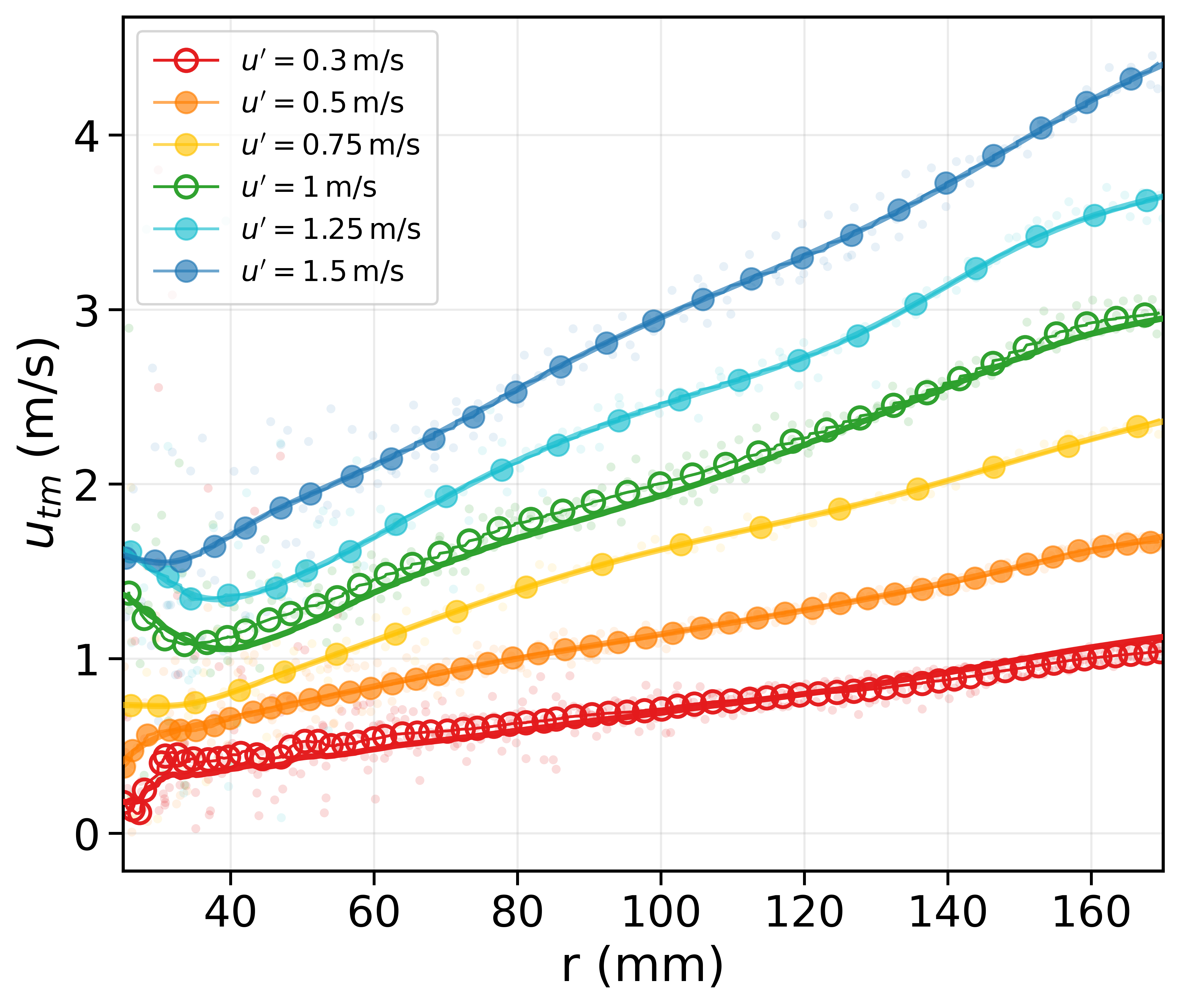}
\put(1,85){\textbf{(b)}}
\end{overpic}
\end{minipage}
\hfill
\begin{minipage}{0.32\textwidth}
\begin{overpic}[width=\linewidth]{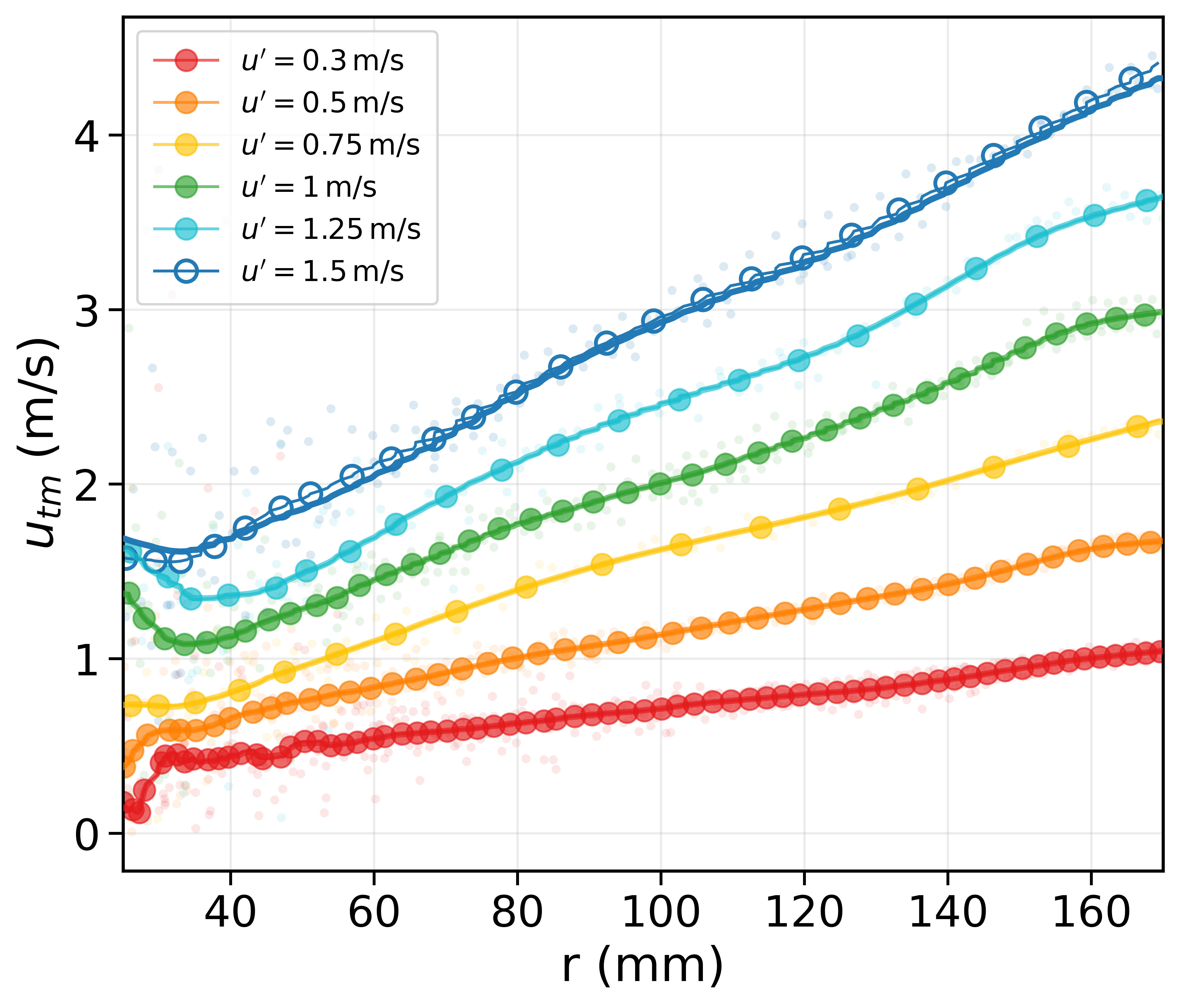}
\put(1,85){\textbf{(c)}}
\end{overpic}
\end{minipage}
\caption{Multi-fidelity prediction of the turbulent mass burning velocity $u_{tm}(r)$ for Case H$_2$-V. Faint scattered symbols represent the raw pressure-derived measurements, while filled circular markers denote the processed (smoothed) high-fidelity data used for supervision. Continuous curves correspond to MuFiNNs predictions. Open markers indicate turbulence intensity levels excluded from training (held-out cases), used to assess interpolation and extrapolation performance. Solid curves denote MuFiNNs predictions. (a) Interpolation test, where two intermediate turbulence intensity levels are excluded from training. (b) Combined interpolation--extrapolation test, where both an intermediate and the lowest turbulence intensity levels are held out. (c) Extrapolation test, where the highest turbulence intensity level is excluded from training.}
\label{fig:H2V_results}
\end{figure*}

For turbulent burning velocity prediction, the MuFiNNs architecture takes the flame radius $r$ and turbulence intensity $u'$ as inputs, and predicts the corresponding turbulent burning velocity $u_{tm}$. The predictive capability of the proposed MuFiNNs framework is evaluated across all considered cases. Representative results for Case H$_2$-V are presented in Fig.~\ref{fig:H2V_results}, which illustrates three training scenarios designed to assess interpolation and extrapolation performance across turbulence intensity levels.

In the first configuration [Fig.~\ref{fig:H2V_results}(a)], two intermediate turbulence intensity levels are excluded from the training dataset to evaluate interpolation capability. In this setting, the MuFiNNs model reconstructs the held-out conditions using only neighboring Hi-Fi data together with the underlying low-fidelity trend. The predicted curves remain in strong agreement with the experimental measurements, indicating that the model effectively captures the continuous dependence of $u_{tm}$ on both $r$ and $u'$.

To evaluate interpolation and extrapolation simultaneously, both an intermediate and the lowest turbulence intensity levels are excluded from the training dataset, as shown in Fig.~\ref{fig:H2V_results}(b). In this case, the model is required to infer missing conditions both within and near the boundary of the training domain. The predictions continue to follow the experimental trends closely, demonstrating robust generalization across partially observed regimes.

The extrapolation capability of the model is examined in Fig.~\ref{fig:H2V_results}(c) by withholding the highest turbulence intensity level during training. This scenario is particularly important from a practical perspective, as experimental measurements at high turbulence intensities are often difficult or costly to obtain. Despite the absence of direct supervision at high $u'$, the MuFiNNs predictions successfully capture both the overall trend and magnitude of $u_{tm}$. Notably, even though the Hi-Fi data exhibit increased scatter and a less well-defined trend—particularly at smaller radii—the model remains stable and produces physically consistent predictions. This demonstrates that the combination of Lo-Fi guidance and nonlinear HF correction enables reliable extrapolation even in regimes with noisy or weakly structured data.

\section{Conclusions}

In this work, a hierarchical multi-fidelity neural network (MuFiNNs) framework was developed for predictive modeling of turbulent premixed flames under limited high-fidelity data availability. By combining sparse experimental measurements with structured low-fidelity representations, the framework captures coupled flame geometry and turbulent burning dynamics while preserving dominant physical trends and correcting nonlinear discrepancies through high-fidelity learning. In particular, the hierarchical formulation proved effective when the dimensionality of the parameter space exceeded what could be supported by the available experimental data, enabling stable learning across thermodynamic and turbulence-dependent operating conditions.

The results demonstrate accurate reconstruction of high-fidelity measurements, reliable interpolation across unseen conditions, and meaningful extrapolation beyond the training domain, including experimentally challenging regimes involving elevated turbulence and pressure. Importantly, the framework remained effective even when trained on noisy or weakly structured data, highlighting the potential of multi-fidelity learning to extract predictive physical behavior from sparse experiments where purely data-driven approaches may be unreliable. More broadly, these findings establish hierarchical multi-fidelity learning as a practical strategy for experimental combustion modeling in data-limited regimes and suggest a pathway for integrating sparse experiments with structured surrogate physics in reactive-flow prediction. At the same time, quantitative extrapolation across entirely unseen thermodynamic combinations remains limited by the sparse coverage of the current dataset, emphasizing the need for broader pressure–temperature sampling in future studies.

Although regression-based approximations were used here to generate low-fidelity representations, the framework is general and can naturally incorporate reduced-order or physics-based models when available, offering a route toward improved predictive fidelity and interpretability. This generality makes the approach promising for more complex reactive-flow configurations and broader scientific machine-learning applications in combustion.

\section*{Declaration of competing interest}
The authors declare no competing interests.

\section*{Acknowledgments}
SZ and SJ would like to acknowledge support from DoD MURI Award N00014-23-1-2499. YX and JY would like to thank EPSRC (Grant No.~EP/W002299/1) for financial support.

% ===================== References ================
\bibliographystyle{cnf-num}
\bibliography{cnf-refs}

@article{Zheng2025CnF,
  title   = {Extracting global reaction rate and turbulent flame speed from reconstructed 3D spherically expanding flames},
  author  = {Zheng, Yutao and Ahmed, Pervez and Hochgreb, Simone},
  journal = {Combustion and Flame},
  volume  = {278},
  pages   = {114247},
  year    = {2025},
  doi     = {10.1016/j.combustflame.2025.114247}
}

@article{Xie2025CnF3D,
  author  = {Xie, Yu and Yang, Junfeng and Ahmed, Pervez and Thorne, Benjamin John Alexander and Gu, Xiaojun},
  title   = {Three-dimensional dynamics of unstable lean premixed hydrogen--air flames: Intrinsic instabilities and morphological characteristics},
  journal = {Combustion and Flame},
  volume  = {271},
  pages   = {113800},
  year    = {2025},
  doi     = {10.1016/j.combustflame.2024.113800}
}

@article{Xie2024CnFwrinkling,
  author  = {Xie, Yu and Yang, Junfeng and Gu, Xiaojun},
  title   = {Flame wrinkling and self-disturbance in cellularly unstable hydrogen--air laminar flames},
  journal = {Combustion and Flame},
  volume  = {265},
  pages   = {113505},
  year    = {2024},
  doi     = {10.1016/j.combustflame.2024.113505}
}

@article{Xie2023FuelPulsation,
  author  = {Xie, Yu and Morsy, Mohamed Elsayed and Yang, Junfeng},
  title   = {Self-acceleration and global pulsation of unstable laminar hydrogen--air flames},
  journal = {Fuel},
  volume  = {353},
  pages   = {129182},
  year    = {2023},
  doi     = {10.1016/j.fuel.2023.129182}
}

@article{Xie2022FuelCellular,
  author  = {Xie, Yu and Morsy, Mohamed Elsayed and Li, Jinzhou and Yang, Junfeng},
  title   = {Intrinsic cellular instabilities of hydrogen laminar outwardly propagating spherical flames},
  journal = {Fuel},
  volume  = {327},
  pages   = {125149},
  year    = {2022},
  doi     = {10.1016/j.fuel.2022.125149}
}

@article{LiXie2024FuelEthaneH2,
  author  = {Li, Jinzhou and Xie, Yu and Morsy, Mohamed Elsayed and Yang, Junfeng},
  title   = {Laminar burning velocities, {Markstein} numbers and cellular instability of spherically propagation ethane/hydrogen/air premixed flames at elevated pressures},
  journal = {Fuel},
  volume  = {364},
  pages   = {131078},
  year    = {2024},
  doi     = {10.1016/j.fuel.2024.131078}
}

@article{Zeldovich1944,
  author  = {Zel'dovich, Ya. B.},
  title   = {On the theory of flame propagation},
  journal = {Zhurnal Eksperimental'noi i Teoreticheskoi Fiziki},
  year    = {1944},
  note    = {Classic early work (in Russian); bibliographic variants exist.}
}

@article{Markstein1951,
  author  = {Markstein, George H.},
  title   = {Nonsteady flame propagation},
  journal = {Journal of the Aeronautical Sciences},
  year    = {1951},
  note    = {Bibliographic details may vary by archive edition.}
}

@article{Clavin1985,
  author  = {Clavin, Paul},
  title   = {Dynamic behavior of premixed flame fronts in laminar and turbulent flows},
  journal = {Progress in Energy and Combustion Science},
  year    = {1985},
  volume  = {11},
  pages   = {1--59}
}

@article{MatalonMatkowsky1982,
  author  = {Matalon, Moshe and Matkowsky, B. J.},
  title   = {Flames as gasdynamic discontinuities. Part {II}. Unsteady flame dynamics},
  journal = {Journal of Fluid Mechanics},
  year    = {1982},
  volume  = {124},
  pages   = {239--259}
}

@article{Sivashinsky1977,
  author  = {Sivashinsky, G. I.},
  title   = {Nonlinear analysis of hydrodynamic instability in laminar flames. Part 1. Derivation of basic equations},
  journal = {Acta Astronautica},
  year    = {1977},
  volume  = {4},
  pages   = {1177--1206}
}

@article{MichelsonSivashinsky1977,
  author  = {Michelson, D. M. and Sivashinsky, G. I.},
  title   = {Nonlinear analysis of hydrodynamic instability in laminar flames. Part 2. Numerical experiments},
  journal = {Acta Astronautica},
  year    = {1977},
  volume  = {4},
  pages   = {1207--1222}
}

@article{BechtoldMatalon2001,
  author  = {Bechtold, J. K. and Matalon, M.},
  title   = {The dynamics of curved premixed flames},
  journal = {Annual Review of Fluid Mechanics},
  year    = {2001},
  note    = {Add volume/pages if needed from your library manager.}
}

@article{BychkovLiberman2000,
  author  = {Bychkov, V. V. and Liberman, M. A.},
  title   = {Dynamics and stability of premixed flames},
  journal = {Physics Reports},
  volume  = {325},
  number  = {4--5},
  pages   = {115--237},
  year    = {2000},
  doi     = {10.1016/S0370-1573(99)00081-2}
}

@book{Law2006,
  author    = {Law, C. K.},
  title     = {Combustion Physics},
  publisher = {Cambridge University Press},
  year      = {2006}
}

@article{Matalon2009,
  author  = {Matalon, M.},
  title   = {Flame dynamics},
  journal = {Proceedings of the Combustion Institute},
  year    = {2009},
  volume  = {32},
  number  = {1},
  pages   = {57--82}
}

@book{WilliamsBook,
  author    = {Williams, Forman A.},
  title     = {Combustion Theory},
  publisher = {Addison-Wesley},
  year      = {1985},
  edition   = {2}
}

@book{Peters2000,
  author    = {Peters, Norbert},
  title     = {Turbulent Combustion},
  publisher = {Cambridge University Press},
  year      = {2000}
}

@book{PoinsotVeynante2005,
  author    = {Poinsot, Thierry and Veynante, Denis},
  title     = {Theoretical and Numerical Combustion},
  publisher = {R. T. Edwards},
  year      = {2005},
  edition   = {2}
}

@article{VeynanteVervisch2002,
  author  = {Veynante, Denis and Vervisch, Luc},
  title   = {Turbulent combustion modeling},
  journal = {Progress in Energy and Combustion Science},
  year    = {2002},
  volume  = {28},
  pages   = {193--266}
}

@article{CandelPoinsot1990,
  author  = {Candel, S. M. and Poinsot, T. J.},
  title   = {Flame stretch and the balance equation for the flame area},
  journal = {Combustion Science and Technology},
  volume  = {70},
  number  = {1--3},
  pages   = {1--15},
  year    = {1990},
  doi     = {10.1080/00102209008951608}
}

@article{Bray1990,
  author  = {Bray, K. N. C.},
  title   = {Studies of the turbulent burning velocity},
  journal = {Proceedings of the Royal Society of London. Series A: Mathematical and Physical Sciences},
  volume  = {431},
  number  = {1882},
  pages   = {315--335},
  year    = {1990}
}

@article{Bradley2003,
  author  = {Bradley, D. and Lau, A. K. C. and Lawes, M. and {others}},
  title   = {Turbulent burning velocities of freely propagating flames in a combustion bomb},
  journal = {Combustion and Flame},
  year    = {2003},
  volume  = {135},
  pages   = {503--523}
}

@article{HowarthAspden2012,
  author  = {Howarth, T. and Aspden, A. J.},
  title   = {A mixing model study of the high-pressure turbulent burning velocity of methane--air flames},
  journal = {Journal of Fluid Mechanics},
  year    = {2012},
  volume  = {701},
  pages   = {87--115}
}

@article{Aspden2011,
  author  = {Aspden, A. J. and Day, M. S. and Bell, J. B.},
  title   = {Turbulence--flame interactions in lean premixed hydrogen: Transition to the distributed burning regime},
  journal = {Journal of Fluid Mechanics},
  volume  = {680},
  pages   = {287--320},
  year    = {2011},
  doi     = {10.1017/jfm.2011.164}
}

@article{ChenIm2018,
  author  = {Chen, J. H. and Im, H. G.},
  title   = {Dynamics of turbulent premixed hydrogen flames with differential diffusion},
  journal = {Combustion and Flame},
  year    = {2018},
  note    = {Representative supporting reference; fill volume/pages if desired.}
}

@article{Ahmed2024PoF,
  author  = {Ahmed, Pervez and Thorne, Benjamin John Alexander and Yang, Junfeng},
  title   = {Development of a multiple laser-sheet imaging technique for the analysis of three-dimensional turbulent explosion flame structures},
  journal = {Physics of Fluids},
  volume  = {36},
  number  = {8},
  pages   = {085112},
  year    = {2024},
  doi     = {10.1063/5.0207937}
}

@article{Ahmed2021,
  author  = {Ahmed, Pervez and Thorne, Benjamin and Lawes, Malcolm and Hochgreb, Simone and Nivarti, Girish V. and Cant, R. Stewart},
  title   = {Three dimensional measurements of surface areas and burning velocities of turbulent spherical flames},
  journal = {Combustion and Flame},
  volume  = {233},
  pages   = {111586},
  year    = {2021},
  doi     = {10.1016/j.combustflame.2021.111586}
}

@article{Forrester2007MF,
  author  = {Forrester, Alexander I. J. and S{\'o}bester, Andr{\'a}s and Keane, Andy J.},
  title   = {Multi-fidelity optimization via surrogate modelling},
  journal = {Proceedings of the Royal Society A},
  year    = {2007},
  volume  = {463},
  number  = {2088},
  pages   = {3251--3269},
  doi     = {10.1098/rspa.2007.1900}
}

@article{Perdikaris2017MF,
  author  = {Perdikaris, Paris and Raissi, Maziar and Damianou, Andreas and Lawrence, Neil D. and Karniadakis, George E.},
  title   = {Nonlinear information fusion algorithms for data-efficient multi-fidelity modelling},
  journal = {Proceedings of the Royal Society A},
  year    = {2017},
  volume  = {473},
  number  = {2198},
  pages   = {20160751},
  doi     = {10.1098/rspa.2016.0751}
}

@article{Raissi2019PINN,
  author  = {Raissi, M. and Perdikaris, P. and Karniadakis, G. E.},
  title   = {Physics-informed neural networks: A deep learning framework for solving forward and inverse problems involving nonlinear partial differential equations},
  journal = {Journal of Computational Physics},
  volume  = {378},
  pages   = {686--707},
  year    = {2019},
  doi     = {10.1016/j.jcp.2018.10.045}
}

@article{Karniadakis2021PIML,
  author  = {Karniadakis, G. E. and Kevrekidis, I. G. and Lu, L. and Perdikaris, P. and Wang, S. and Yang, L.},
  title   = {Physics-informed machine learning},
  journal = {Nature Reviews Physics},
  volume  = {3},
  number  = {6},
  pages   = {422--440},
  year    = {2021},
  doi     = {10.1038/s42254-021-00314-5}
}

@article{Raissi2020HiddenFluid,
  author  = {Raissi, M. and Yazdani, A. and Karniadakis, G. E.},
  title   = {Hidden fluid mechanics: Learning velocity and pressure fields from flow visualizations},
  journal = {Science},
  volume  = {367},
  number  = {6481},
  pages   = {1026--1030},
  year    = {2020},
  doi     = {10.1126/science.aaw4741}
}

@article{Raissi2018DeepHiddenPhysics,
  author  = {Raissi, M.},
  title   = {Deep hidden physics models: Deep learning of nonlinear partial differential equations},
  journal = {Journal of Machine Learning Research},
  volume  = {19},
  number  = {25},
  pages   = {1--24},
  year    = {2018},
  url     = {http://jmlr.org/papers/v19/18-046.html}
}

@article{Dabiri2023RhINN,
  title={Fractional rheology-informed neural networks for data-driven identification of viscoelastic constitutive models},
  author={Dabiri, Donya and Saadat, Milad and Mangal, Deepak and Jamali, Safa},
  journal={Rheologica Acta},
  volume={62},
  number={10},
  pages={557--568},
  year={2023},
  publisher={Springer},
  doi={10.1007/s00397-023-01430-0}
}

@article{Dabiri2025fPINNReview,
  title={A detailed and comprehensive account of fractional Physics-Informed Neural Networks: From implementation to efficiency},
  author={Dabiri, Donya and DaRosa, Joshua and Saadat, Milad and Mangal, Deepak and Jamali, Safa},
  journal={arXiv preprint arXiv:2506.11241},
  year={2025}
}

@article{saberi2025rheoformer,
  author  = {Saberi, Maedeh and Farimani, Amir Barati and Jamali, Safa},
  title   = {RheOFormer: A generative transformer model for simulation of complex fluids and flows},
  journal = {arXiv preprint arXiv:2510.01365},
  year    = {2025},
  doi     = {10.48550/arXiv.2510.01365}
}

@article{Mahmoudabadbozchelou2022nnPINNs,
  title={nn-PINNs: Non-Newtonian physics-informed neural networks for complex fluid modeling},
  author={Mahmoudabadbozchelou, Mohammadamin and Karniadakis, George Em and Jamali, Safa},
  journal={Soft Matter},
  volume={18},
  pages={172--185},
  year={2022},
  publisher={Royal Society of Chemistry},
  doi={10.1039/d1sm01298c}
}

@article{Zolfaghari2026NonlocalPINN,
  author    = {Saghar Zolfaghari and Safa Jamali},
  title     = {Non-local physics-informed neural networks for forward and inverse solutions of granular flows},
  journal   = {arXiv preprint arXiv:2602.16081},
  year      = {2026},
  doi       = {10.48550/arXiv.2602.16081},
  url       = {https://arxiv.org/abs/2602.16081}
}

@article{Zhang2016RandomizedNN,
  author  = {Zhang, Lin and Suganthan, Ponnuthurai N.},
  title   = {A survey of randomized algorithms for training neural networks},
  journal = {Information Sciences},
  volume  = {364--365},
  pages   = {146--155},
  year    = {2016},
  doi     = {10.1016/j.ins.2016.01.039}
}

@article{Saadat2024MFNN,
  author  = {Saadat, Milad and Hartt, William H. and Wagner, Norman J. and Jamali, Safa},
  title   = {Data-driven constitutive meta-modeling of nonlinear rheology via multifidelity neural networks},
  journal = {Journal of Rheology},
  volume  = {68},
  number  = {5},
  pages   = {679--693},
  year    = {2024},
  doi     = {10.1122/8.0000831}
}

@article{Buhrmester2021,
  author  = {Buhrmester, Vanessa and M{\"u}nch, David and Arens, Michael},
  title   = {Analysis of explainers of black box deep neural networks for computer vision: A survey},
  journal = {Machine Learning and Knowledge Extraction},
  volume  = {3},
  number  = {4},
  pages   = {966--989},
  year    = {2021},
  doi     = {10.3390/make3040048}
}

@article{Carvalho2019,
  author  = {Carvalho, Diogo V. and Pereira, Eduardo M. and Cardoso, Jaime S.},
  title   = {Machine learning interpretability: A survey on methods and metrics},
  journal = {Electronics},
  volume  = {8},
  number  = {8},
  pages   = {832},
  year    = {2019},
  doi     = {10.3390/electronics8080832}
}

@article{Lennon2023,
  author  = {Lennon, Kyle R. and McKinley, Gareth H. and Swan, James W.},
  title   = {Scientific machine learning for modeling and simulating complex fluids},
  journal = {Proceedings of the National Academy of Sciences},
  volume  = {120},
  number  = {17},
  pages   = {e2304669120},
  year    = {2023},
  doi     = {10.1073/pnas.2304669120}
}

@article{Meng2019PPINN,
  author  = {Meng, Xuhui and Li, Zhiping and Zhang, Di and Karniadakis, George Em},
  title   = {PPINN: Parareal physics-informed neural network for time-dependent PDEs},
  journal = {arXiv preprint arXiv:1909.10145},
  year    = {2019}
}

@article{Meng2020MPINN,
  author  = {Meng, Xuhui and Karniadakis, George Em},
  title   = {A composite neural network that learns from multi-fidelity data: Application to function approximation and inverse PDE problems},
  journal = {Journal of Computational Physics},
  volume  = {401},
  pages   = {109020},
  year    = {2020},
  doi     = {10.1016/j.jcp.2019.109020}
}

@article{Pang2019fPINN,
  author  = {Pang, Guofei and Lu, Lu and Karniadakis, George Em},
  title   = {fPINNs: Fractional physics-informed neural networks},
  journal = {SIAM Journal on Scientific Computing},
  volume  = {41},
  number  = {4},
  pages   = {A2603--A2626},
  year    = {2019},
  doi     = {10.1137/18M1229845}
}

@article{Pang2020nPINN,
  author  = {Pang, Guofei and D'Elia, Marta and Parks, Michael and Karniadakis, George Em},
  title   = {nPINNs: Nonlocal physics-informed neural networks for a parametrized nonlocal universal Laplacian operator},
  journal = {arXiv preprint arXiv:2004.04276},
  year    = {2020}
}

@article{Lu2019DeepONet,
  author  = {Lu, Lu and Jin, Pengzhan and Karniadakis, George Em},
  title   = {DeepONet: Learning nonlinear operators for identifying differential equations based on the universal approximation theorem of operators},
  journal = {arXiv preprint arXiv:1910.03193},
  year    = {2019}
}

@article{Mangal2025,
  author  = {Mangal, Deepak and Saadat, Milad and Jamali, Safa},
  title   = {Learning a family of rheological constitutive models using neural operators},
  journal = {Journal of Rheology},
  volume  = {69},
  number  = {2},
  pages   = {55--67},
  year    = {2025},
  doi     = {10.1122/8.0000908}
}

@article{Mahmoudabadbozchelou2021MFNN,
  author  = {Mahmoudabadbozchelou, Mohammadamin and Caggioni, Marco and Shahsavari, Setareh and Hartt, William H. and Karniadakis, George Em and Jamali, Safa},
  title   = {Data-driven physics-informed constitutive metamodeling of complex fluids: A multifidelity neural network ({MFNN}) framework},
  journal = {Journal of Rheology},
  volume  = {65},
  number  = {2},
  pages   = {179--198},
  year    = {2021},
  doi     = {10.1122/8.0000138}
}

@article{Mahmoudabadbozchelou2022DigitalRheometerTwins,
  author  = {Mahmoudabadbozchelou, Mohammadamin and Kamani, Krutarth M. and Rogers, Simon A. and Jamali, Safa},
  title   = {Digital rheometer twins: Learning the hidden rheology of complex fluids through rheology-informed graph neural networks},
  journal = {Proceedings of the National Academy of Sciences of the United States of America},
  volume  = {119},
  number  = {20},
  pages   = {e2202234119},
  year    = {2022},
  doi     = {10.1073/pnas.2202234119}
}

\end{document}